\documentclass[lettersize,journal]{IEEEtran}
\usepackage{amsmath,amssymb,amsfonts}
\usepackage{algpseudocode}
\usepackage{algorithm}
\usepackage{array}
\usepackage{textcomp}
\usepackage{stfloats}
\usepackage{url}
\usepackage{verbatim}
\usepackage{graphicx}
\usepackage[font=footnotesize]{caption}
\usepackage{subcaption}
\usepackage{cite}
\usepackage{xcolor}
\usepackage[frozencache,cachedir=.]{minted}

\usepackage{url}
\usepackage[detect-none,binary-units=true]{siunitx}
\usepackage{tabularx,booktabs}
\usepackage{hhline}
\usepackage{datetime2}
\usepackage{multirow}

\usepackage{epstopdf}

\usepackage{pgf,tikz}
\usetikzlibrary{arrows}

\usepackage{colortbl}
\usepackage{soul}
\usepackage{pifont}
\usepackage{color}
\usepackage{alltt}
\usepackage{enumerate}
\usepackage{pbox}
\usepackage{eurosym}
\usepackage{authblk}
\usepackage{cleveref}

\usepackage{todonotes}

\newcommand{\mAh}{mAh}
\newcommand{\mps}{\meter / \second}

\definecolor{ethblue}{RGB}{0,105,180}
\definecolor{ethgrey}{RGB}{111,111,110}
\definecolor{colpbl}{HTML}{000579}
\definecolor{colcstb}{HTML}{FFC009}
\definecolor{colost}{HTML}{FF6F00}
\definecolor{colcsmm}{HTML}{CE4289}
\definecolor{colemcc}{HTML}{E32C2D}
\definecolor{colyellow}{HTML}{E3E300}

\begin{document}

\title{Robust and Efficient Depth-based Obstacle Avoidance for Autonomous Miniaturized UAVs
}

\author{Hanna M\"uller, \IEEEmembership{Student Member, IEEE}, Vlad Niculescu, \IEEEmembership{Student Member, IEEE}, Tommaso Polonelli, \IEEEmembership{Member, IEEE}, Michele Magno, \IEEEmembership{Senior Member, IEEE} and Luca Benini, \IEEEmembership{Fellow, IEEE}

\thanks{\today, Manuscript created June, 2022; 
}}



\makeatletter
\newcommand\footnoteref[1]{\protected@xdef\@thefnmark{\ref{#1}}\@footnotemark}
\makeatother
\maketitle

\begin{abstract}
Nano-size drones hold enormous potential to explore unknown and complex environments. Their small size makes them agile and safe for operation close to humans and allows them to navigate through narrow spaces. However, their tiny size and payload restrict the possibilities for on-board computation and sensing, making fully autonomous flight extremely challenging. The first step towards full autonomy is reliable obstacle avoidance, which has proven to be technically challenging by itself in a generic indoor environment. Current approaches utilize vision-based or 1-dimensional sensors to support nano-drone perception algorithms. This work presents a lightweight obstacle avoidance system based on a novel millimeter form factor 64 pixels multi-zone Time-of-Flight (ToF) sensor and a generalized model-free control policy. Reported in-field tests are based on the Crazyflie 2.1, extended by a custom multi-zone ToF deck, featuring a total flight mass of \SI{35}{\gram}. The algorithm only uses 0.3\% of the on-board processing power (\SI{210}{\micro\second} execution time) with a frame rate of \SI{15}{fps}, providing an excellent foundation for many future applications.
Less than 10\% of the total drone power is needed to operate the proposed perception system, including both lifting and operating the sensor. The presented autonomous nano-size drone reaches 100\% reliability at \SI{0.5}{\mps} in a generic and previously unexplored indoor environment. The proposed system is released open-source with an extensive dataset including ToF and gray-scale camera data, coupled with UAV position ground truth from motion capture.  
\end{abstract}

\begin{IEEEkeywords}
UAV, Autonomous navigation, Nano-drones, Perception, Obstacle Avoidance, ToF Array
\end{IEEEkeywords}

\section{Introduction}

Unmanned aerial vehicles (UAVs) are nowadays used for monitoring, inspection, surveillance, transportation, logistics and many other fields \cite{shakhatreh2019unmanned}. In several scenarios, a small form factor brings advantages - smaller drones are more agile and can traverse complex environments ranging from cluttered offices to industrial facilities, allowing safe operation close to humans in locations otherwise inaccessible \cite{gyagenda2022review,miiller2021funfiiber}. Nano-UAVs \cite{ps2020mini} that weigh a few tens of grams mostly rely on off-board computation due to highly restricted on-board capabilities, typically milliwatt-power microcontrollers (MCU), strongly limited by power and size constraints \cite{miiller2021funfiiber}. MCUs are not powerful enough to run state-of-the-art solutions such as complex navigation models and simultaneous localization and mapping (SLAM)~\cite{song2021autonomous, loquercio2021learning}. On the other hand, relying only on on-board sensing and computation brings many advantages - higher reliability when wireless links fail (or are jammed), lower latency in control actions, and reduced bandwidth requirements if external control is limited to high-level commands. Until now, autonomous exploration with the same agility and safety as an expert human pilot has been confined to low speed due to the lack of compact integrated low-power sensors, and resource-constrained navigation strategies \cite{gyagenda2022review, song2021autonomous}.

The major challenge for nano-UAVs is achieving autonomous navigation through a reliable and universal obstacle avoidance policy and trajectory planning in real-world applications \cite{rezwan2022artificial}. 
To enable on-board decisions based on the nano-UAV surroundings, the processing should use only a minor fraction, i.e., 10\%, of the overall energy envelope. For instance, in nano-UAV platforms like the Crazyflie 2.1, in which the total power is around \SI{10}{\watt} including also the motors, the maximum processing power needs to be in the orders of hundreds of mW to do not substantially affect the flying time \cite{elkunchwar2021toward}. This power budget is compatible with a simple MCU, such as a general-purpose ARM Cortex-M4 core, commonly used on nano-UAVs, featuring a clock speed of just a few hundred \SI{}{\mega\hertz} and just $\sim$\SI{200}{\kilo\byte} of RAM.

The motivation to enable local and lightweight navigation policies on highly constrained platforms pushes the research to explore alternative solutions that are not a direct downscale of their bigger and more powerful counterparts~\cite{mcguire2019comparative}, reconsidering the whole perception pipeline, from the sensor to the navigation strategy \cite{mcguire2019minimal, coppola2020survey}. Moreover, the on-board pipeline has to be robust against disturbances, such as sensor noise, illumination conditions and motion blur \cite{loquercio2021learning}.
Obstacle avoidance is commonly vision-based, exploiting deep neural network (DNN) approaches to extract the navigation context from the scene~\cite{niculescu2021improving,loquercio2021learning,wang2020uav}, but also solutions with laser range finders or even radars exist~\cite{schouten2019biomimetic,yasin2020unmanned,zhao20203, duisterhof2019learning}. On nano-UAVs, approaches with cameras, mono or stereo, have been investigated \cite{bahnam2021stereo, niculescu2021improving}. However, they suffer from a fundamental drawback of a high number of input pixels and, therefore, a high computational load, which can absorb a significant fraction of the computational capabilities of an MCU, even a powerful multicore one~\cite{garofalo2020pulp}. Moreover, they are dataset-dependent, meaning that in general, the on-board network needs to be retrained if the operational environment changes \cite{niculescu2021improving,loquercio2018dronet,li2020visual}. This effect is further exacerbated by the limited amount of on-board memory, pushing researchers to optimize model size at the cost of decreased generalization.

This work focuses on infrastructure less and autonomous exploration, where the nano-UAV can move through an unknown environment without a local/remote infrastructure for supporting positioning or distributed perception. 
Existing algorithms often rely on global map planning with known obstacles or are not suitable for position-blind contexts \cite{wang2019autonomous}. Moreover, they offer only a limited set of local decisions, like the \textit{wall following} approach in the bug algorithm literature that generally exploits single point laser-beam sensors \cite{mcguire2019comparative}. Single-beam rangers have the drawback of only returning a one dimensional measurement, making it impossible to acquire a depth map with a single sensor at a high frequency. Placing multiple sensors or using previously available multi-zone sensors is too power-hungry and heavy for operation on nano-drones. Engineering a more complex policy than wall following can be challenging, requiring tailored heuristics to find dataset-dependent architectures \cite{loquercio2021learning, coppola2020survey}, motivating the researchers to implement model-free methods \cite{mcguire2019minimal}, which have the potential to ease the engineering process, providing also a robust solution.    

This work proposes a complete depth-based perception system optimized for lightweight and low-power flying robots and targeted to support obstacle avoidance and enable autonomous indoor navigation of nano-UAVs. Our solution exploits a novel commercial multi-zone ranger to automatically extracts complex obstacle geometry from the background. The decision policy, together with the preprocessing, is a model-free solution; it replaces sophisticated frameworks targeted at high-end platforms with an algorithm that fits commercial MCU specs. We employed the VL53LC5CX, an 8x8 or 4x4 pixel ToF sensor from STMicroelectronics, which can generate a depth map and a pixel validity matrix at zero-computational cost and a frame rate up to \SI{60}{fps}. We empirically characterized the VL53LC5CX field of view (FoV), which ranges between \SI{20}{\centi\meter} and \SI{4}{\meter}. The pixel validity indicator simplifies filtering outliers or out-of-range measurements, resulting in highly computationally efficient navigation for a low latency response. 

As a reference platform, we selected the Crazyflie 2.1 from Bitcraze, on which a \SI{2.49}{\gram} custom expansion board was designed to support two VL53LC5CX in opposite directions, front and back-facing w.r.t the flying direction. The obstacle avoidance policy, developed on the reported dataset, consists of a decision tree fed by the depth 8x8 matrix. It runs in real-time, with a frame processing time of just \SI{210}{\micro\second}, on the Crazyflie MCU, an ARM Cortex-M4f, using a mere 0.31\% of its computational capacity. The depth information, pre-filtered by removing invalid pixels, is categorized into four zones to control the 3D spatial movements of the nano-UAV and the flying speed. 

Although our perception solution requires only a fraction (9.4\%) of the total power budget, results show best-in-class performance, with virtually zero crash rate flying at \SI{0.5}{\mps}, the possibility to avoid moving obstacles at up to \SI{2}{\mps} and to randomly explore complex unknown environments characterized by narrow pipes (\SI{65}{\centi\meter}) and thin reflective objects with standard ambient light and in complete darkness. Our lightweight model-free perception system successfully demonstrates its potential in real-world experiments characterized by outdoor and indoor environments other than controlled mazes. 

On average, the speed of \SI{1}{\mps} appears to be the best balance between crashing probability and flight distance, respectively below 20\% and \SI{100}{\meter} in a variety of complex real case studies. The main scientific contributions of this work are listed below. 
(\textit{i}) We designed a lightweight (\SI{2.49}{\gram}) perception board for nano-UAVs with up two VL53LC5CX sensors. With the Crazyflie 2.1 it can be used as a plug \& play expansion board. 
(\textit{ii}) We leverage a compact integrated multi-region (8x8) ToF sensor to extract depth information without the support of standard vision-based frameworks and complex model-based approaches. 
(\textit{iii}) We empirically characterized the ToF sensor in nano-UAV flying conditions, demonstrating the possibility of using this SoC solution to extract precise and reliable depth information from the scene. 
(\textit{iv}) We collected a dataset containing 43 records showing time-synchronized data from a grey-scale CMOS camera, one depth matrix extracted from the VL53LC5CX, the internal Crazyflie state, and an absolute 3D position measured by a mocap system. 
(\textit{v}) We developed a lightweight obstacle avoidance and random exploration algorithm that can be executed in real-time on a resource-constrained microcontroller. It interactively reacts to complex obstacle geometries, commanding the escape maneuver to the internal state controller. It is based on a model-free decision tree that groups objects at different distances and locations, which is easily extendable by adding path planning and mapping capabilities.
(\textit{vi}) We carried out real-world assessment and performance evaluation in multiple operating conditions, such as mazes, indoor/outdoor environments, narrow passages, and moving obstacles. Evaluation metrics are based on the maximum flight velocity, cumulative and individual success rate, flying time, distance, perception artifacts and latency and, lastly, the flight trajectory. 
(\textit{vii}) The whole project, together with the hardware, the dataset, and the obstacle avoidance policy, is released open-source\footnote{\label{note:github}\url{https://github.com/ETH-PBL/Matrix_ToF_Drones}}.
\section{Related work}

UAVs are massively adopted in real-life scenarios, from civilian applications \cite{mohamed2020unmanned} -- such as surveillance,
transportation, environmental and industrial monitoring, agriculture services, and first aid -- to military services \cite{shakeri2019design}.
In particular, indoor navigation enables smart buildings and drone-machine or drone-human interaction, opening new research frontiers \cite{polonelli2020flexible}.
In this area, nano-drones have great potential \cite{rezwan2022artificial}, which is why nowadays there are numerous research projects aiming to address open challenges for enabling autonomous nano-UAV navigation, mapping, automation, distributed computing and swarm formations~\cite{shakhatreh2019unmanned,rezwan2022artificial,miiller2021funfiiber,mohamed2020unmanned}.

\begin{table*}[ht]
\caption{Comparison against SoA works on perception-based navigation.}
\label{tab:soa-comparison}
\centering
\renewcommand{\arraystretch}{1.2}
\begin{tabular}{c c c c c c c c c c}
\hline
\hline
\begin{tabular}[c]{@{}c@{}} Reference work \end{tabular} & Model-free & Vision-based & \begin{tabular}[c]{@{}c@{}}Fully\\ on-board\end{tabular} & \begin{tabular}[c]{@{}c@{}}Moving\\ obstacles\end{tabular} & \begin{tabular}[c]{@{}c@{}}In-field\\ evaluation\end{tabular} & \begin{tabular}[c]{@{}c@{}}Maximum\\ speed [m/s]\end{tabular} & \begin{tabular}[c]{@{}c@{}}Maximum\\ flight time [s]\end{tabular} & \begin{tabular}[c]{@{}c@{}}Dataset \\ release\end{tabular} & \begin{tabular}[c]{@{}c@{}}Code available\end{tabular} \\ \hline
Our work                                               & \checkmark         & 	$\times$           & \checkmark                                                      & \checkmark                                                         & \checkmark                                                            & 2.66                                                          & 443                                                               & \checkmark                                                         & \checkmark                                                        \\ 
\cite{niculescu2021improving}                                                    & 	$\times$          & \checkmark           & \checkmark                                                      & 	\checkmark                                                          & \checkmark                                                            & 2.29                                                          & 216                                                               & 	\checkmark                                                          & \checkmark                                                        \\ 
\cite{mcguire2019minimal}                                                & 	$\times$          & 	$\times$            & \checkmark                                                      & 	$\times$                                                          & \checkmark                                                            & N/S                                                           & N/S                                                               & 	$\times$                                                          & 	$\times$                                                         \\ 
\cite{duisterhof2021tiny}                                             & 	$\times$          & 	$\times$            & \checkmark                                                      & 	$\times$                                                          & \checkmark                                                            & 1.0                                                             & 200                                                              & 	$\times$                                                          & 	$\times$                                                         \\ 
\cite{li2020visual}                                         & \checkmark         & \checkmark           & \checkmark                                                      & 	$\times$                                                          & \checkmark                                                            & 2.6                                                           & N/S                                                               & 	$\times$                                                          & 	$\times$                                                         \\ 
\cite{chathurangasensor}                                                  & \checkmark         & 	$\times$            & 	$\times$                                                       & 	$\times$                                                          & \checkmark                                                            & 0.4                                                           & N/S                                                               & 	$\times$                                                          & 	$\times$                                                         \\ \hline
\hline
\end{tabular}
\end{table*}
%

State-of-the-art exploration solutions that proved to work well on conventional drones, such as SLAM, are still too resource-demanding for nano-UAVs~\cite{niculescu2021improving,foehn2021alphapilot}.
While other works proved that mapping is also possible with nano-drones~\cite{chathurangasensor}, their approach relies on off-board computing, which introduces the need of having a computer in the loop, limited range, and communication overhead.

Perceiving the 2-D/3-D structure of the environment is vital for the functionality of UAVs or robotic systems in general~\cite{loquercio2021learning}, as it enables path planning and autonomous navigation through mapping and obstacle avoidance~\cite{rovira2022review}. 
High-end UAV platforms extract the 3-D environmental information relying on complex, specialized neural networks. 
By sensing the environment and interacting with other agents~\cite{rovira2022review}, they learn to generate a depth map estimating distances to objects using a variety of active sensors like monocular and stereo cameras~\cite{loquercio2021learning, schilling2019learning}, asynchronous event cameras~\cite{gehrig2021combining}, structured light~\cite{muglikar2021event}, lidar~\cite{yasuda2020autonomous}, and ToF sensors~\cite{mcguire2019minimal} sampling the scene at a fixed scan rate.

Scaramuzza \textit{et. al} demonstrated the possibility to fly at high speed in complex environments, such as forests, exploiting a stereo camera and a neural network trained only on synthetic data sets \cite{loquercio2021learning}. 
To remove the context bias from the simulator environment, and then train the algorithms to work on a generalized scenarios, the authors do not directly process RGB images, but a depth map is extracted from the Intel RealSense 435 stereo pairs. 
Using the depth matrix not only for obstacle avoidance but also for mapping stages, they achieved a maximum fly speed of \SI{10}{\mps} in real scenarios, in which the drones featured a success rate above 50\% - 100\% below \SI{8}{\mps} - in various and unknown environments.

In \cite{loquercio2021learning}, the authors present a state-of-the-art approach to enable autonomous navigation in indoor and outdoor environments, based on depth estimation. 
However, their methodology cannot be applied on nano-UAV platforms as it demands high memory (i.e., gigabytes) and computational requirements.
Furthermore, their approach also relies on high-resolution sensing, which is an uncrossable technological barrier for nano-UAVs \cite{miiller2021funfiiber}, leaving \textit{de facto} the nano-UAV autonomous exploration still an open challenge.

Due to recent technological advancements, miniature depth sensors are becoming a reality, being already incorporated in commercial devices such as smartphones or top-range quadrotors. 
The SONY DepthSense IMX556PLR back-illuminated ToF image sensor\footnote{\url{www.sony-depthsensing.com}} features a resolution of 640 x 480 pixels with up to \SI{8.3}{\meter} working distance, while the TeraRanger Evo 64px\footnote{\url{www.terabee.com}} proposed a compact 64 pixel and \SI{12}{\gram} solution for robotic application.
The aforementioned commercial sensors feature depth extraction and filtering, directly providing a pixel-by-pixel confidence flag on the sensor, moving relevant computation effort from the computing core. However, they are still not compatible with nano-UAVs platforms due to their weight or incompatible digital interfaces with a commercial MCU, e.g., the IMX556PLR is targeted for high-end cellular processors. Nevertheless, there is a clear research trend in this area, which aims to replace the traditional control framework by using an optimized solution able to sense and extract the depth map with a single SoC commercial component, simplifying the mapping and planning processing latency. 
Moreover, this increases the system's robustness against unexplored flying scenarios, in which dataset-based solutions show limitations or are more error-prone \cite{niculescu2021improving,yasuda2020autonomous}.

\Cref{tab:soa-comparison} presents a comparison between the most recent SoA works on perception-based navigation with nano-drones.
In \cite{niculescu2021improving}, the authors present an automatic deployment flow of a convolutional neural network (CNN) that runs on-board a nano-drone and enables autonomous navigation and obstacle avoidance capabilities. 
However, despite its effectiveness with static and dynamic obstacles, the general performance seems poorer in unfamiliar environments (i.e., not present in the training dataset).
Furthermore, the CNN can reliably detect the presence of an obstacle and reduce the drone's forward velocity, or it can adjust the drone's heading when following a lane. 
However, due to a dataset limitation, it is often unable to steer around an unknown obstacle to avoid collision, especially in narrow corridors. 
In contrast to our approach, where all algorithms run in a single SoC (i.e., STM32), their system takes advantage of an additional multi-core SoC which is in charge of running the CNN and transmitting the inference result to the main MCU, increasing the total mass by \SI{4.4}{\gram}.

The authors of~\cite{mcguire2019minimal} introduce a swarm gradient bug algorithm (SGBA) for enabling autonomous exploration relying on an array of four ToF sensors.
However, the drones can only follow walls and do not perform any localization or surrounding detection. 
The proposed scenario is to find "victims" in an office environment.
However, the video data is stored on SD-cards and has to be read off-board after the flying mission, which introduces a delay in obtaining the information about the environment.
Furthermore, despite the excellent measurement accuracy of the ToF sensors (i.e., 3\% of the full scale), a vision-based solution was necessary to compensate for the poor spatial coverage of the four distance "pixels" at low distances~\cite{mcguire2019minimal}.

Similarly, \cite{duisterhof2021tiny} relies on four ToF sensors and a light sensor and presents an approach based on deep reinforcement learning that enables a nano-drone to seek and find a light source while avoiding obstacles.
However, they do not provide any results on dealing with dynamic obstacles, and the maximum speed they report during the testing phase is \SI{1}{\meter / \second}, which is significantly slower than our system.

The work in \cite{li2020visual} proposes a vision-based system for drone racing, whose goal is to detect "gates" and fly through them as fast as possible.
While effective in passing through gates, their system is tuned to work with a particular type of obstacle and does not deal with general objects in the trajectory.
Although their algorithms run entirely on-board, they use a power-hungry SoC (i.e., Cortex A7 plus a dual-core GPU), resulting in a drone that weighs twice as much as our solution.

Lastly, \cite{chathurangasensor} demonstrates the capabilities of creating a 2-D map of the environment, relying on a particle filter that fuses information from 12 ToF sensors.
They create a custom deck to accommodate the 12 sensors, which they use in combination with a nano-drone, but the whole computation necessary to run their algorithms is carried off-board.

\section{Background and Hardware platform}
This work presents a complete system description of an obstacle avoidance system for nano-UAVs, from the hardware design to in-field evaluations. In this application scenario, design optimization and weight minimization are essential to enable longer flight times. We used the commercial Crazyflie 2.1 platform from Bitcraze, extending its functionality with a custom expansion board and new sensors, such as the VL53L5CX from STMicroelectronics. All used components are commercially available, and our design is released as open-source.
\subsection{Crazyflie}
The Crazyflie 2.1, henceforth Crazyflie, is an open software/hardware nano-UAV commonly used in research. It comes with a base board featuring an inertial measurement unit (IMU), a barometer, radio communication (using an nRF51822 from Nordic Semiconductors), and as the main processor, an STM32F405 (168MHz, 196kB RAM), responsible for sensor readout, state estimation and real-time control. One important feature of the Crazyflie is its extension headers - there is a wide variety of commercially available decks to plug onto the base board to sense the environment and improve state estimation or even plan where to fly. We use a downward-facing \textit{Flow-deck v2}, featuring an optical flow sensor and a 1D ToF sensor to improve the position estimation computed by the extended Kalman Filter (eKF). To collect the dataset, we also connected an \textit{AI-deck}, featuring a QVGA greyscale camera and WiFi to stream the images to a local computer.
It features a Himax HM01B0, an ultra-low-power 320×240 grayscale camera with a \SI{115}{\degree} diagonal FoV and a NINA-W102 WiFi module from U-Blox. 
In our application, the  8+1 core RISC-V MCU does not directly communicate with the STM32F405, processing and compressing in a parallel task the acquired frame, which is then sent to a local gateway through a WiFi link and then timestamped at the arrival together with the Crazyflie state transmitted over Bluetooth (nRF51822).  
The base version of the Crazyflie weighs \SI{27}{\gram} and can fly up to 7 minutes with its \SI{250}{\mAh} battery; adding the \textit{Flow-deck v2} adds \SI{1.6}{\gram} and the \textit{AI-deck} another \SI{4.4}{\gram}. 
The maximum payload that still enables take-off is \SI{15}{\gram}. 
However, the maneuverability and flight time are very poor when flying with the maximum payload~\cite{elkunchwar2021toward}. 
To increase the flight time with multiple connected decks, we use a \SI{350}{\mAh} battery instead of the \SI{250}{\mAh} one that comes with the commercial drone, which features a 30C current rate to support high motor current peaks but adds \SI{1.1}{\gram} of extra-payload. 
In total, a maximum of \SI{7.9}{\gram} are available for further decks.
\subsection{ToF multi-zone sensor}
The VL53L5CX\footnote{\label{note1}\url{www.st.com/en/imaging-and-photonics-solutions/vl53l5cx.html}} is designed for a wide range of ambient lighting conditions, and it is based on a vertical cavity surface emitting diode (VCSEL), a single-photon avalanche diode (SPAD) array, physical infrared filters, and diffractive optical elements (DOE).
The novel feature of the VL53L5CX is the multi-zone capability; it can provide a matrix of either 8x8 or 4x4 pixels configurable by software. Each zone provides a distance measurement, and in case of ToF miss-calculation or interference at \SI{940}{\nano\meter} light-wave, an error flag is reported. This way, noise and errors can be filtered out through a validity matrix overlapped with the measurement matrix. From \SI{2}{\centi\meter} to \SI{2}{\meter} the ranging accuracy is characterized by STMicroelectronics as an absolute value ($\pm$\SI{15}{\milli\meter}), above \SI{2}{\meter} the overall ranging accuracy degrades up to 11\% of the absolute distance\footnoteref{note1}, with a working range of up to \SI{4}{\meter}. 

The VL53L5CX can be configured in different ranging modes, with varying integration times, resolutions, ranging frequencies and sharpener values.
There are two ranging modes: continuous ranging and autonomous ranging - in continuous ranging, the VCSEL is always on and therefore, the integration time is maximized, while in autonomous mode, the integration time can be configured, saving energy by turning off the VCSEL when not used. Two different resolutions are available, either 4x4 pixel or 8x8 pixels. The maximal ranging frequency is dependent on the resolution; for 4x4 pixels \SI{60}{\hertz} can be reached; for 8x8 pixels the limit is \SI{15}{\hertz}. 

\begin{figure}[t]
  \centering
  \includegraphics[width=0.97\linewidth]{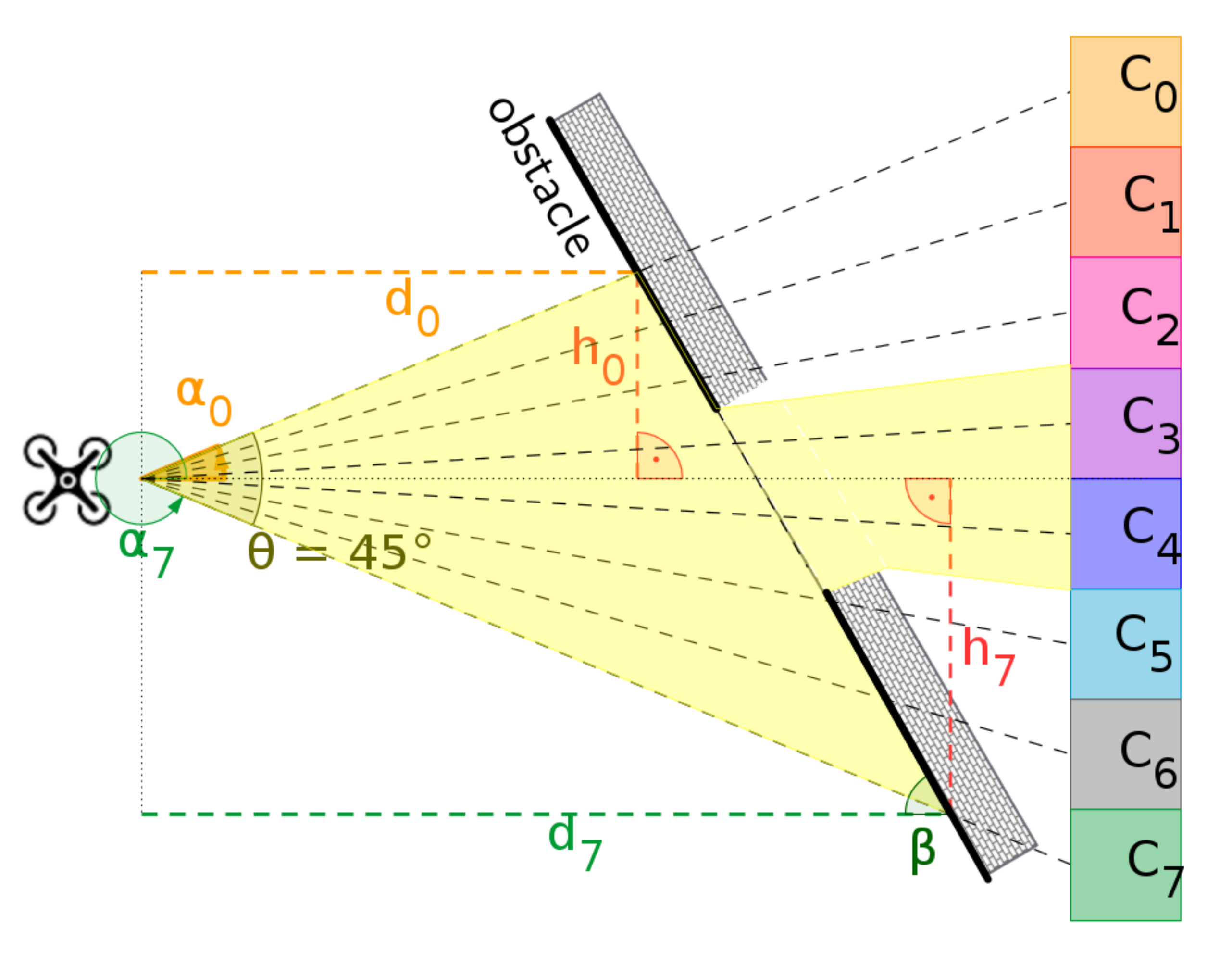}
      \caption{The drone faces an obstacle with a gap (e.g. a door) with an angle $\beta$. $C_x$ is the corresponding column associated with the 8x8 matrix, while $d_x$ is the projects planar distance. The term $h_x$ is calculated using the ToF sensor FoV and the measured $d_x$.}
  \label{fig:droneangle}
\end{figure}

As the returned signal from a target does not have sharp edges, the sharpener value can be configured to remove some of the signal caused by veiling glare\footnoteref{note1}.
The FoV depends on the environment (target distance and reflectance, ambient light level) and the sensor configuration (resolution, ranging mode, integration time, sharpener). To ensure proper functionality, the cover window opening has to be at least as wide as the exclusion zone (\SI{61}{\degree} vertically and \SI{55.5}{\degree} horizontally). However, the detection volume\footnoteref{note1} is narrower than the exclusion zone; it is reduced to around \SI{45}{\degree}. Figure~\ref{fig:droneangle} visualizes the functionality, showing the drone facing an angled ($\beta$) wall with a gap (e.g. a door). The multi-zone ToF sensor measures $d_x$, but as angle $\alpha_x$ is known from the FoV, $h_x$ can be computed.
To interface the VL53L5CX with a commercial MCU, a standard \SI{400}{\kilo\hertz} I2C digital bus is required, along with two GPIOs. Both are available on the STM32F405. The power supply spans between \SI{2.8}{\volt} and \SI{3.3}{\volt}, thus making it compatible with most of the MCUs and the open-source nano-UAV platforms available on the market.
\subsection{Multi-zone ranger deck}
\begin{figure}[t]
    \centering
    \includegraphics[width=0.9\columnwidth]{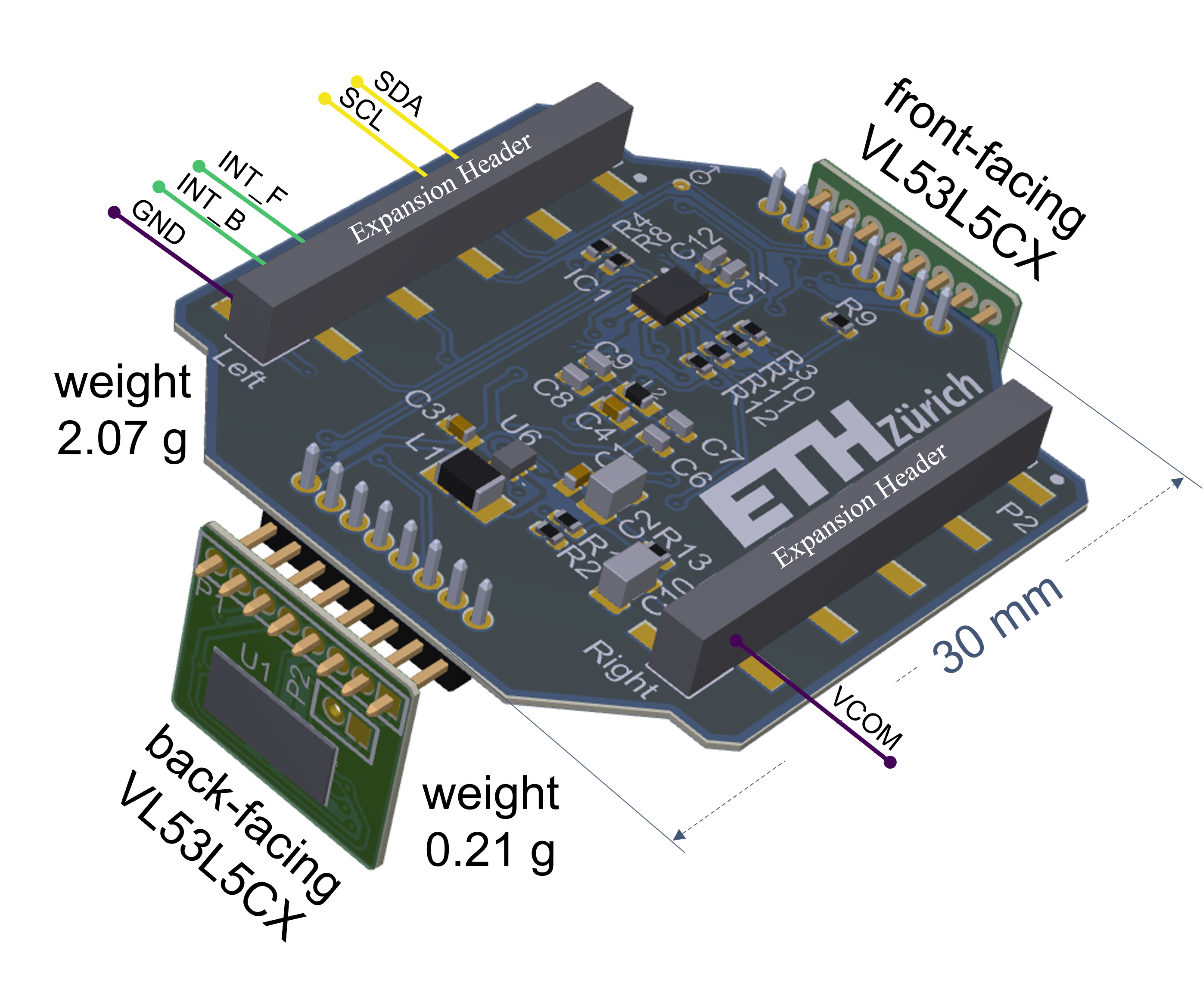}
    \caption{The open source multi-zone ToF deck compatible with the Crazyflie 2.1. A forward and a backward facing VL53L5CX can be mounted vertically to a base board. The maximum weight is \SI{2.49}{\gram} with a size of \SI{9}{\centi\metre\squared}.}
    \label{fig:cf_tof_deck}
\end{figure}

\begin{figure}[t]
    \centering
    \begin{subfigure}[t]{1.0\columnwidth}
        \centering
        \includegraphics[width=\columnwidth]{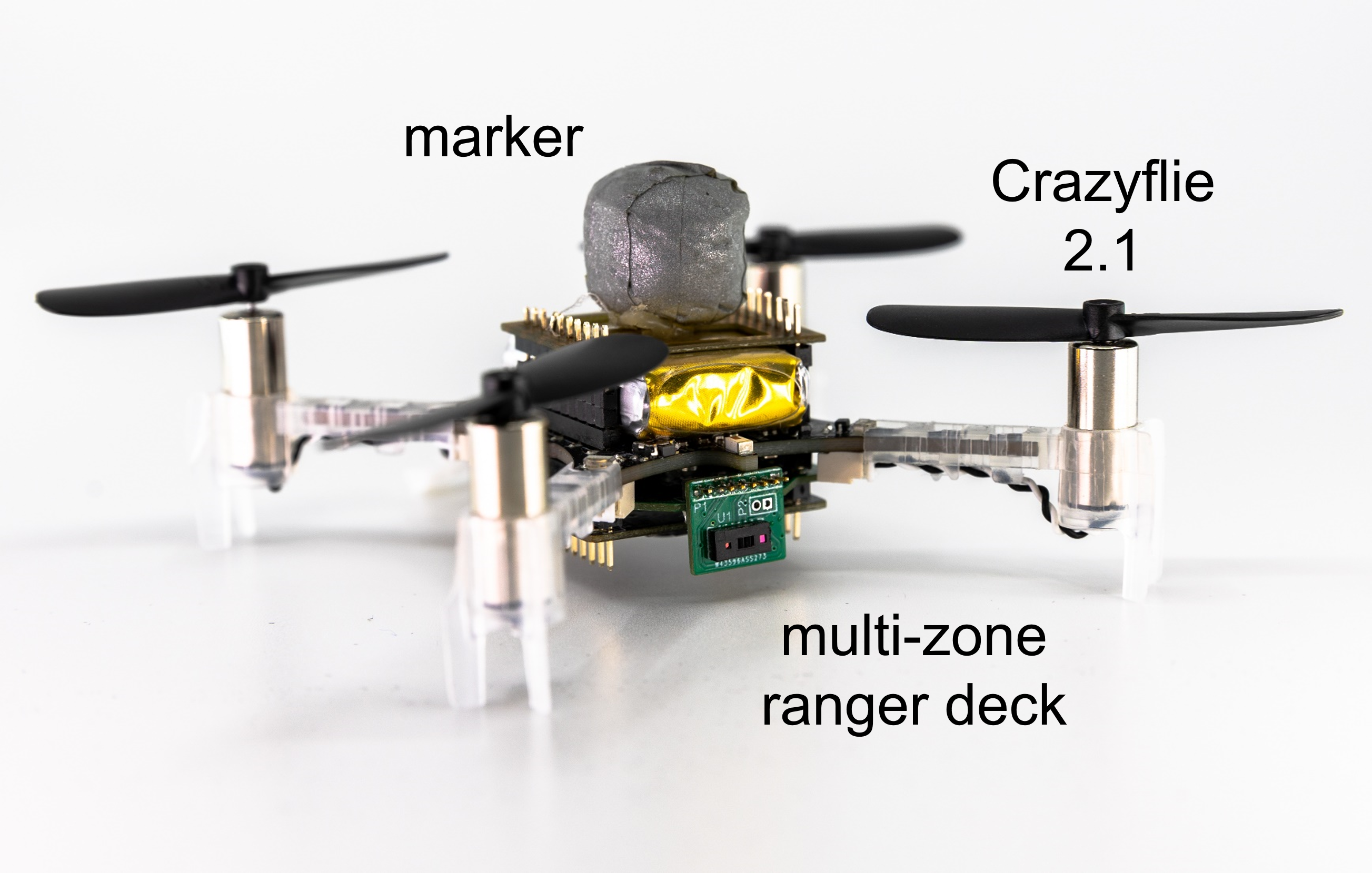}
        \caption{Hardware setup featuring a flow deck and our custom multi-zone ToF deck on the bottom of the Crazyflie, combined with a battery holder and a Vicon marker on top.}
        \label{fig:cf_tof}
    \end{subfigure}
    \hfill
    \centering
    \begin{subfigure}[t]{1.\columnwidth}
        \centering
        \includegraphics[width=\columnwidth]{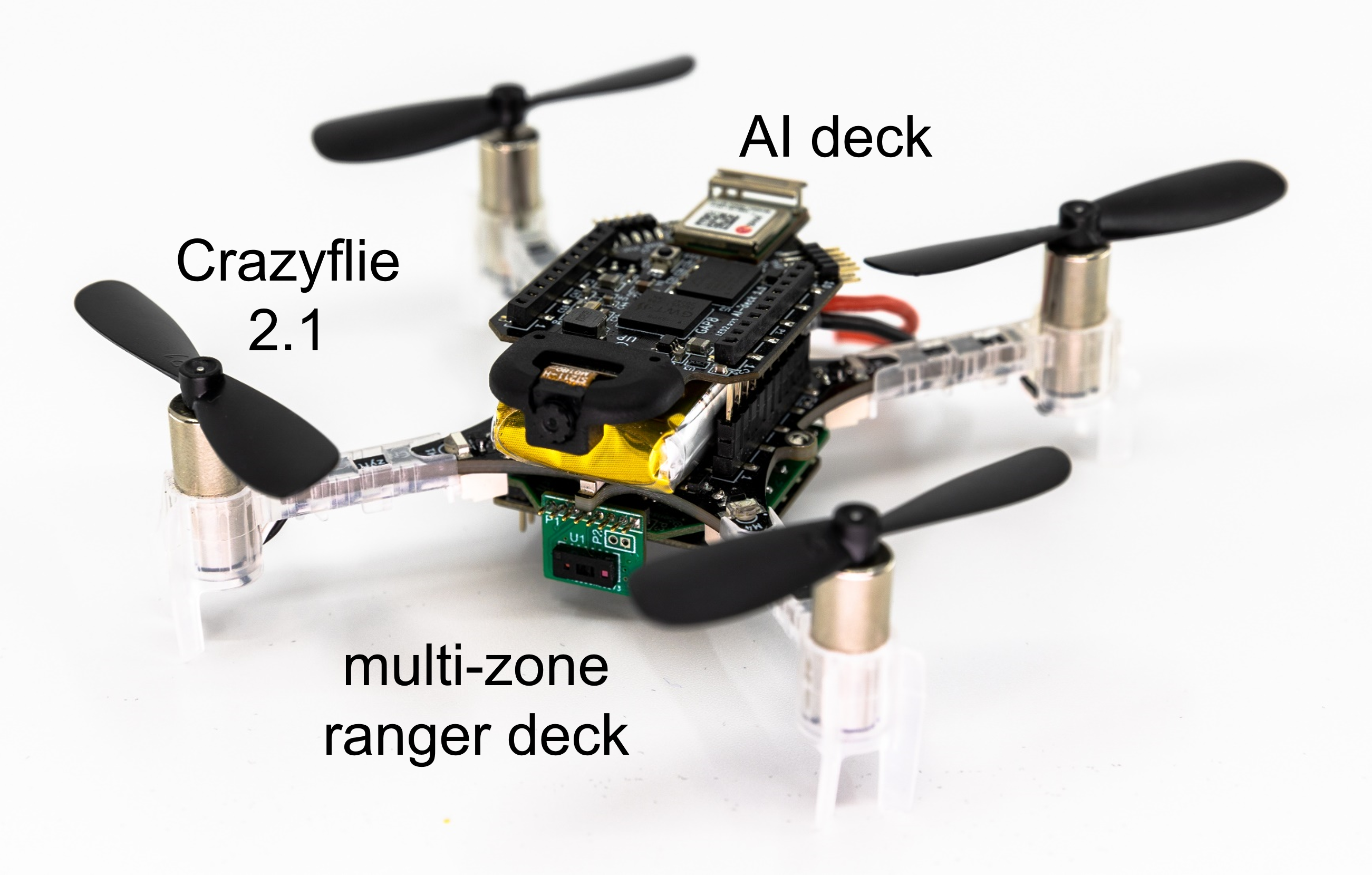}
        \caption{For the dataset collection, the AI-deck was added instead of the battery holder. In this configuration an extra \SI{4.4}{\gram} payload is added.}
        \label{fig:cf_tof_ai}
    \end{subfigure}
    \caption{Our hardware setup for data collection and in-field testing.}
    \label{fig:cf_tof_tot}
\end{figure}

To support in-field tests and complex flight paths, exploiting the full Crazyflie performances, we designed a custom deck specifically optimized for the VL53L5CX ToF sensor. Our new multi-zone ranger deck, shown in Figure~\ref{fig:cf_tof_deck}, can be used at the same time as the \textit{AI-deck} and the \textit{Flow-deck v2}. The multi-zone ranger deck features two mounting positions for a VL53L5CX sensor, one in the front and one in the back, enabling the possibility to detect obstacles in the front or the back. For this paper, we investigate the flying performances using only the forward VL53L5CX, as shown in Figure~\ref{fig:cf_tof}. Each sensor requires \SI{286}{\milli\watt} in continuous acquisition mode; thus, for providing a stable and low noise \SI{3}{\volt} power source, we use the TPS62233 step-down switching voltage regulator with the battery voltage as input. The TCA6408A - I2C GPIO expander manages the reset and power-down pins to decrease the amount of used line on the Crazyflie connector and to ensure compatibility with the \textit{AI-deck} and the \textit{Flow-deck v2}. 
Independent interrupt lines are used for each sensor to decrease the frame acquisition latency. The final design, in \Cref{fig:cf_tof_deck}, has a total size of \SI{29.4}{\milli\meter} x \SI{30}{\milli\meter} x \SI{9.5}{\milli\meter} and, in our configuration, a weight of only \SI{2.07}{\gram}. Mounting the back-facing sensor board would add \SI{0.21}{\gram}. As proposed in \Cref{fig:cf_tof}, the final flying setup used in this work uses a multi-zone ranger deck, a \textit{Flow-deck v2}, and a battery holder with a reference marker for performance analysis. The payload is \SI{4.8}{\gram}, whereas for the dataset collection the \textit{AI-deck} adds an extra \SI{4.4}{\gram}. In general, we always mounted the multi-zone ranger deck below the Crazyflie frame (\Cref{fig:cf_tof}) and the \textit{AI-deck} above the battery (\Cref{fig:cf_tof_ai}).
The hardware design, as well as the bill of materials, are released open-source on GitHub\footnoteref{note:github}.
\section{Characterization and Calibration} \label{sec:characterization}
This section provides an empirical evaluation of the sensor to assess its effectiveness in measuring the distance to various objects in different lighting and flying conditions.
Throughout this evaluation, the sensor is mounted at the height of \SI{1}{\meter} from the ground on static support (i.e., a tripod). 
The whole setup is positioned such that the sensor is parallel to a wall, and the whole area covered in the sensor's field of view is flat.
Using this setup, we sweep our sensor within the range of \SI{0.2}{\meter} - \SI{3}{\meter} from the wall while maintaining its orientation.
The movement is performed with a step of \SI{0.2}{\meter}, and at each step, we acquire 1000 distance frames from the sensor using the 8x8 configuration.
In addition to the distance matrix, we also store the measurement validity matrix provided by the sensor, which reports which entries in the distance matrix are trustful. We repeat this acquisition procedure for the following four configurations: \textit{i)} white wall, ambient indoor light (i.e., $\sim$ \SI{500}{\lux}) \textit{ii)} white wall, darkness (i.e., $<$ \SI{10}{\lux}) \textit{iii)} white wall, ambient indoor light \textit{iv)} white wall, darkness. 
The data stored for these scenarios represent the foundation of our characterization.

\begin{figure}[t]
  \centering
  \includegraphics[width=\linewidth]{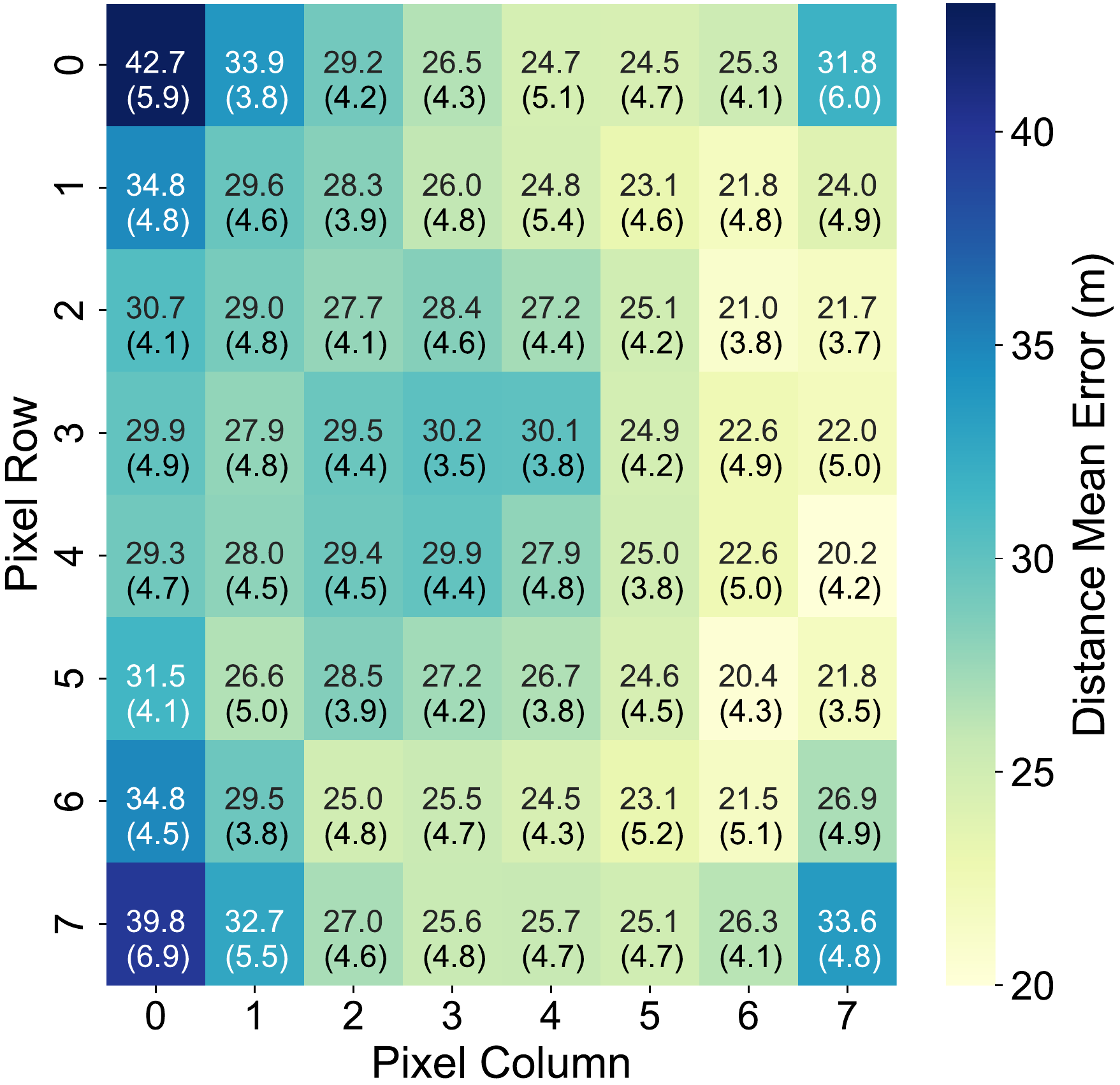}
  \caption{VL53L5CX pixel-by-pixel characterization at \SI{1}{\meter}. Values are in mm. Each pixel includes the offset, on the top, and the variance, bottom, computed over 1000 successive samples in a fixed position.}
  \label{fig:heatmap}
\end{figure}

First, we evaluate the error of the distance measurements in terms of mean and standard deviation. \Cref{fig:heatmap} shows these metrics for the case of a white background with normal ambient light when the sensor is positioned \SI{1}{\meter} far from the wall. 
The error statistics are computed over 1000 samples for each individual pixel in the matrix. 
We note that the highest mean errors in the corners, being about \SI{1}{\centi\meter} -- \SI{2}{\centi\meter} higher than errors associated with the rest of the pixels. 
The mean error takes values in the range of \SI{19}{\milli\meter} -- \SI{42}{\milli \meter} and we remark a gradient in the mean error from left to right, which is most likely because of the imperfections in the sensor alignment. 
The standard deviation of the distance error takes values in the range of \SI{3.4}{\milli \meter} -- \SI{7.3}{\milli \meter}, while the highest values are again encountered in the corners.

\begin{figure}[t]
  \centering
  \includegraphics[width=\linewidth]{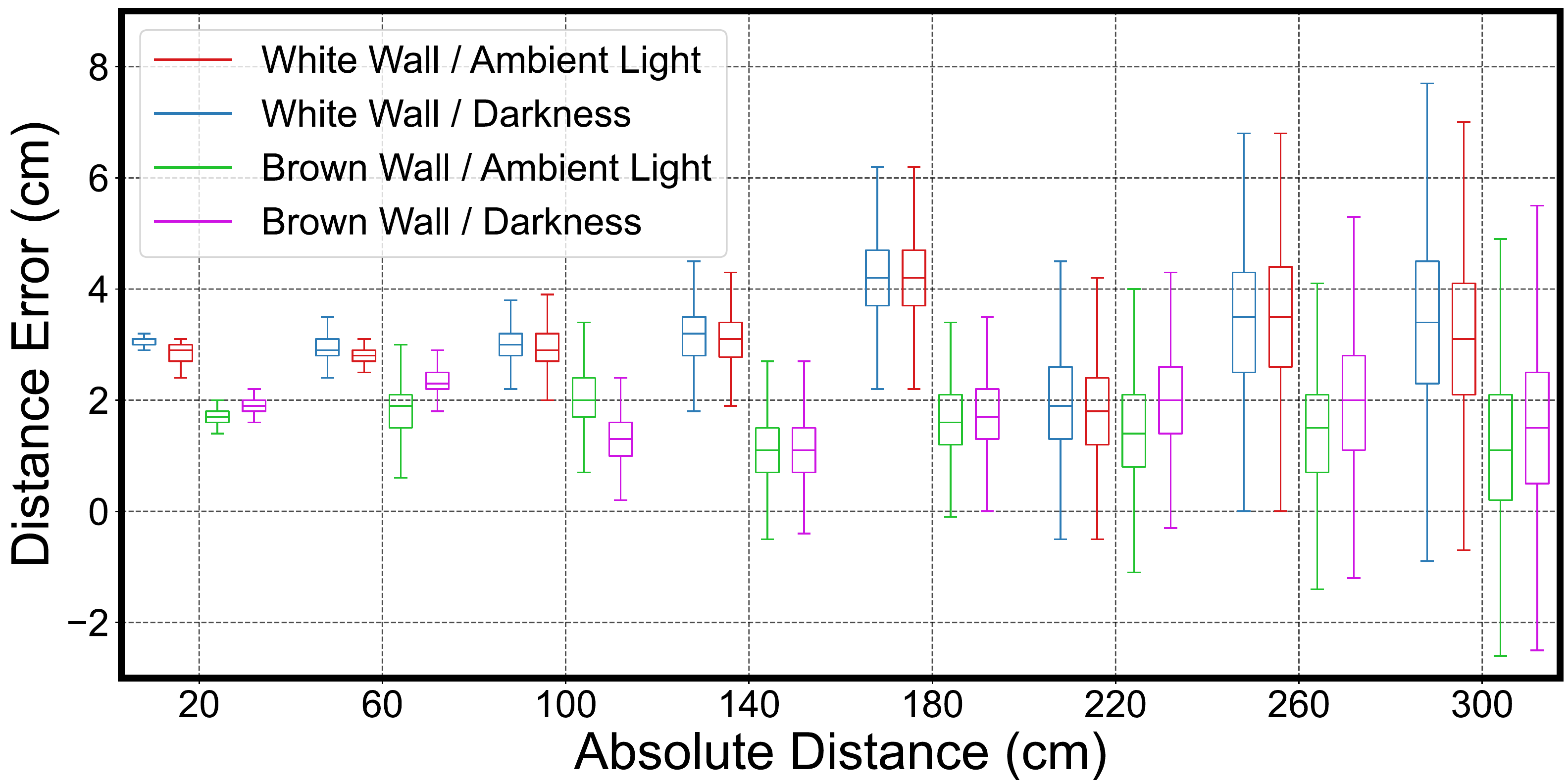}
  \caption{The distance measurement error as a function of the absolute distance. The evaluation is performed for an absolute distance in the range \SI{20}{\centi\meter} --  \SI{300}{\centi\meter} with a step of  \SI{40}{\centi\meter} for each of the four considered scenarios.}
  \label{fig:boxplot}
\end{figure}

Second, we extend the previous investigation to analyze the statistics of the distance error in all the four configurations introduced at the beginning of this section.
Figure~\ref{fig:boxplot} shows the distance error as a function of the distance to the wall for each scenario, considering only one pixel of the matrix (i.e., one of the four inner pixels).
We highlight a pairwise similarity (i.e., median error $<$ \SI{0.5}{\centi\meter}) between the white wall scenarios.
Furthermore, the same pattern applies to the brown wall scenarios, which in terms of median error, seem to lead to better results than the white wall case.
However, this difference seems to be a constant offset, while the min -– max range and inter-quartile difference are about the same for each scenario at a given distance.
The min -- max range spans up to \SI{1}{\centi\meter} for an absolute distance of \SI{20}{\centi\meter} and up to \SI{8}{\centi\meter} for an absolute distance of \SI{3}{\meter}.

\begin{figure}[t]
  \centering
  \includegraphics[width=\linewidth]{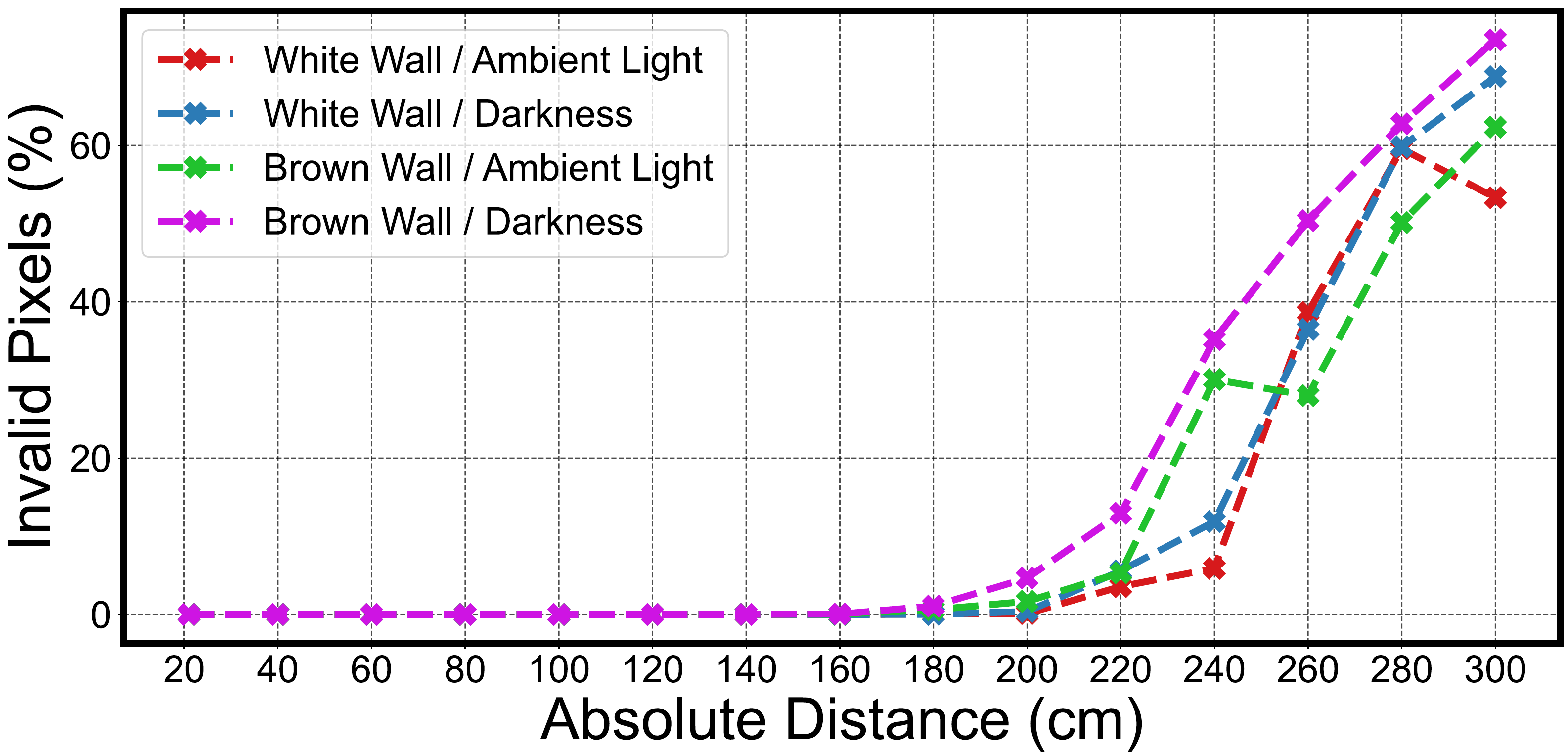}
  \caption{The pixel validity as a function the absolute distance. The evaluation is performed for an absolute distance in the range \SI{20}{\centi\meter} --  \SI{300}{\centi\meter} with a step of  \SI{20}{\centi\meter}. In all the four considered scenarios, the pixel validity decreases when the absolute distance increases.}
  \label{fig:pixel-loss}
\end{figure}

Lastly, we evaluate how the sensor measurement validity decreases with the absolute distance.
The validity depends on the amount of reflected light but also external disturbances, such as ambient lighting. 
Figure~\ref{fig:pixel-loss} shows the measurement validity curves for each of the four considered scenarios.
We point out that the validity is higher than 95\% in all scenarios, given an absolute distance to the wall smaller than \SI{2}{\meter} and higher than 50\% given an absolute distance of \SI{2.6}{\meter}.
Overall, the measurement validity is higher for the scenarios with a white wall due to a higher surface reflectivity.

To conclude the sensor characterization, we can claim that the sensor does not require any calibration phase, as its accuracy is very good within the operating range that we target (i.e., a few meters).
Furthermore, we also observed that the reliability is high for absolute distances of up to \SI{2}{\meter}, which is sufficient for enabling obstacle avoidance on nano-drones.

\section{Dataset}

After static tests and the VL53L5CX empirical assessment, a dynamic dataset was collected in different configurations while maneuvering in indoor environments. Tests were performed in controlled and open spaces, with the support of a motion capture system (mocap) \textit{Vicon Vero 2.2}\footnote{https://www.vicon.com/hardware/cameras/vero/} at a rate of \SI{50}{\hertz}. A human pilot manually steered the Crazyflie. Initially, the dataset was used to develop and test the obstacle avoidance algorithm presented in Section~\ref{sec:algorithm}. However, other researchers can also use it to improve our system by integrating the multi-zone ToF data with processed information from a CNN and the grayscale camera~\cite{niculescu2021improving} or by applying a more general DNN algorithm to enhance on-board intelligence~\cite{liu2022adaptive}. For this reason, we release the acquired data as open source\footnoteref{note:github}.
We collected (\textit{a}) internal state estimation (attitude, velocity, position) of the Crazyflie, (\textit{b}) multi-zone ToF array in 8x8 pixel configuration, (\textit{c}) camera images (QVGA greyscale), (\textit{d}) Vicon data (attitude, position) in a time series format with a millisecond accuracy. The dataset consists of three main groups: object approach moving the drone on a single axis, yaw rotations around the Z-axis, and a general-purpose set of flying tests approaching various obstacles and narrow holes. 
The first dataset group, named \textit{Linear Movements}, consists of 10 recordings of flights with (\textit{a}), (\textit{b}), (\textit{c}), and (\textit{d}) data, approaching a wood panel at different speeds and stopping and flying back always on the same axis, rotations and altitude variations are disabled. The total test time is \SI{216}{\second} with an average of \SI{22}{\second} per acquisition. The next group, \textit{Yaw Rotations}, consists of 3 recordings with (\textit{a}), (\textit{b}), (\textit{c}), and (\textit{d}) data, rotating on a single axis (yaw) at \SI{1}{\meter} from an obstacle. Recorded data reach a total of \SI{94}{\second}. The third and final group, named \textit{Obstacle Avoidance} is composed of 30 recordings with a mixed combination of (\textit{a}), (\textit{b}), (\textit{c}), (\textit{d}) - 14 acquisitions - and (\textit{a}), (\textit{b}), (\textit{c}) - 16 acquisitions. In total, for the third group, \SI{17}{\minute} of flight maneuvers are present in the GitHub\footnoteref{note:github} repository, with an average of \SI{35}{\second} per acquisition. 
\begin{figure}[t]
    \centering
    \begin{subfigure}[t]{0.45\textwidth}
        \centering
        \includegraphics[width=0.8\columnwidth]{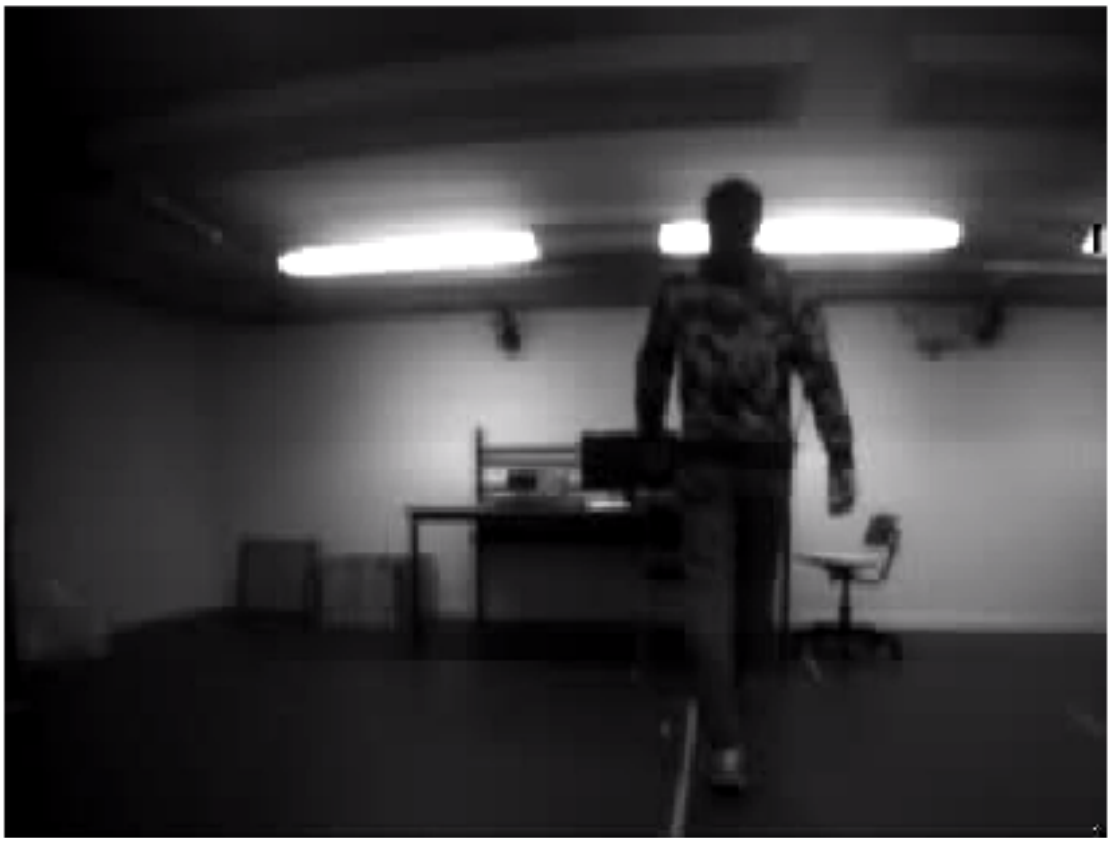}
        \caption{Grayscale QVGA image captured by the \textit{AI-Deck}. A person is walking at approximately 1.7 meter from the nano-drone. Note that the Himax HM01B0 camera is saturating due to the indoor lighting}
        \label{fig:dataset_man_img}
    \end{subfigure}
    \hfill
    \centering
    \begin{subfigure}[t]{0.45\textwidth}
        \centering
        \includegraphics[width=0.8\columnwidth]{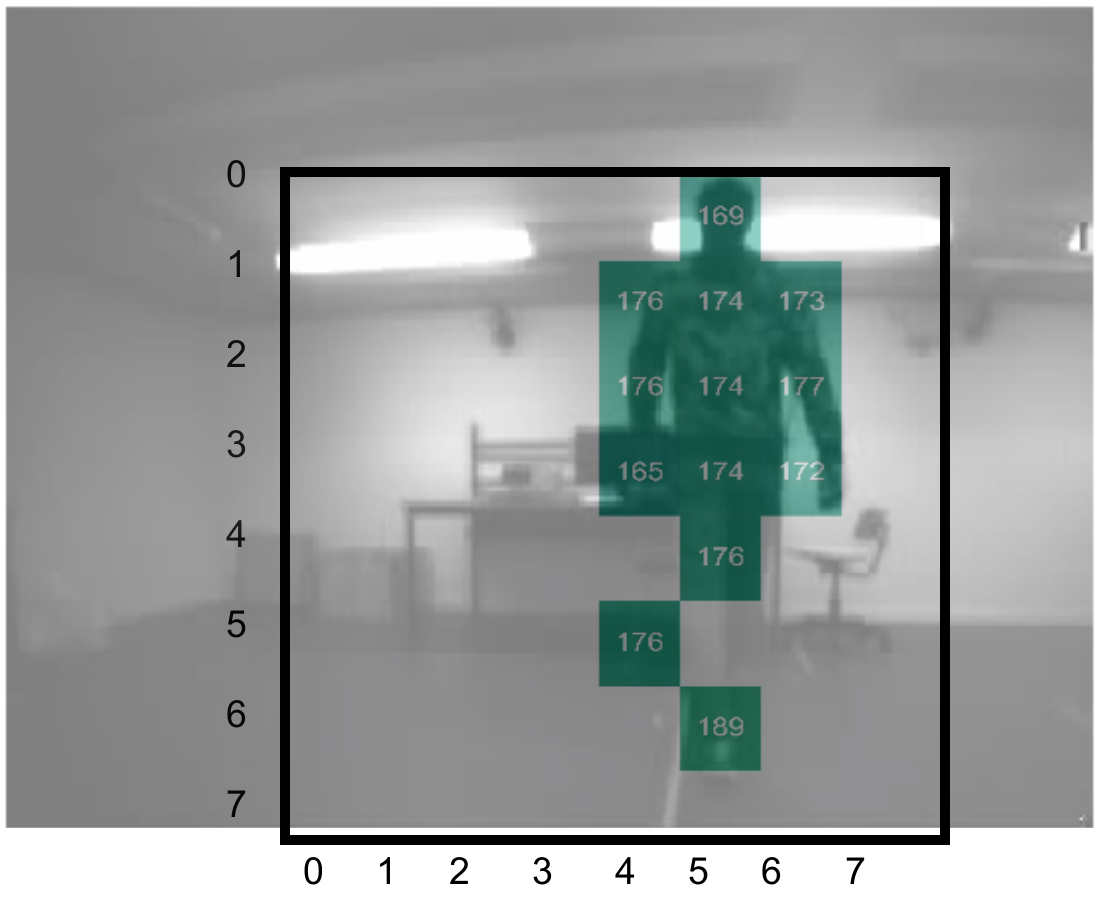}
        \caption{Pre-processed 8x8 depth map from the front-facing multi-zone ToF sensor, where invalid pixels are filtered and not shown. Each pixel features a distance expressed in centimeter and a colormap to help the reader in visualizing the depth of the front facing obstacle. The camera image is displayed as well for visualizing the FoV.}
        \label{fig:dataset_man_tof}
    \end{subfigure}
    \caption{Still scene from the \textit{O16} recording\footnoteref{note:github}. In this test, the UAV took off and hovered, a person walked several times perpendicularly to the drone FoV.  The two figures shows the identical Crazyflie status from different perspective, respectively from the left to the right. (a) The grayscale camera from the \textit{AI-Deck}; (b) Pre-processed 8x8 depth map from the multi-zone ToF sensor, scene equivalent to (a). Note that the Himax HM01B0 and the VL53L5CX have a different FoV and different mounting position on the Crazyflie 2.1 frame - see \Cref{fig:cf_tof_ai} for reference. The camera image is displayed as well for visualizing the FoV.}
    \label{fig:dataset_man}
\end{figure}
\begin{figure}[t]
    \centering
    \begin{subfigure}[t]{0.45\textwidth}
        \centering
        \includegraphics[width=0.8\columnwidth, ]{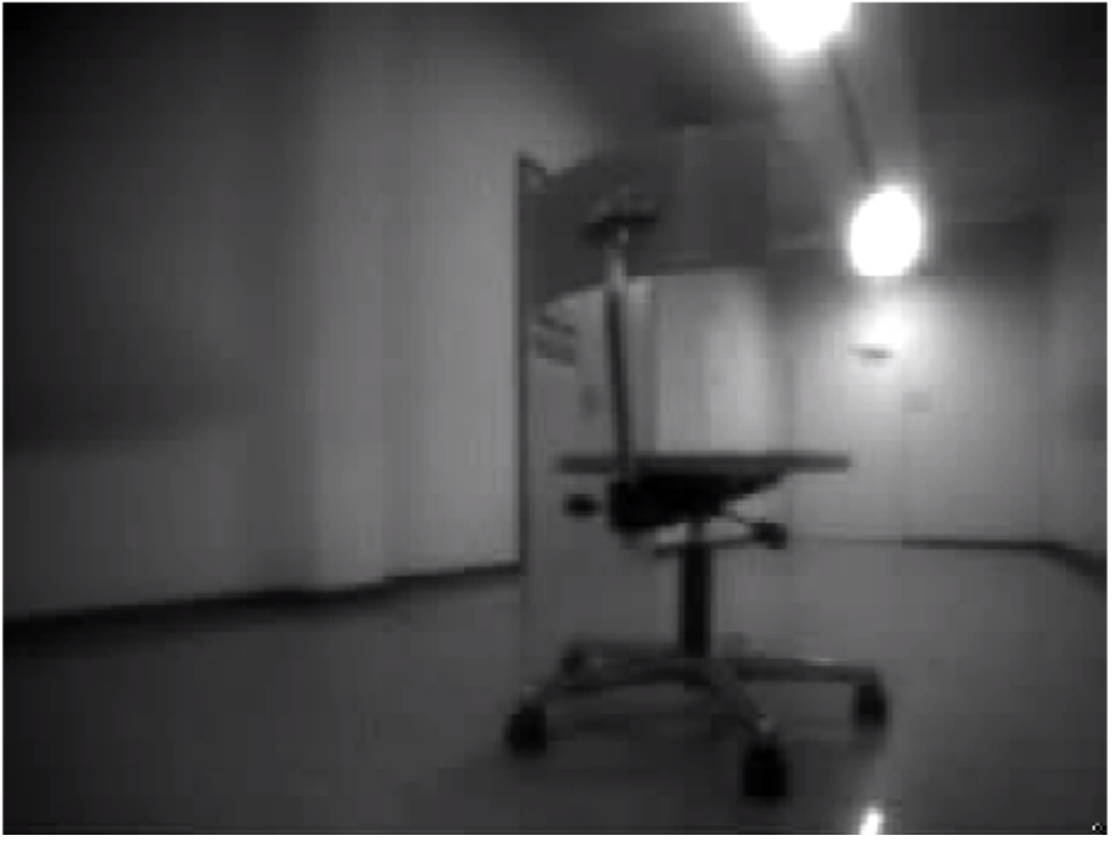}
        \caption{Grayscale QVGA image captured by the \textit{AI-Deck}. A chair is placed at approximately \SI{55}{\centi\meter} from the nano-drone}
        \label{fig:dataset_chair_img}
    \end{subfigure}
    \hfill
    \centering
    \begin{subfigure}[t]{0.45\textwidth}
        \centering
        \includegraphics[width=0.8\columnwidth]{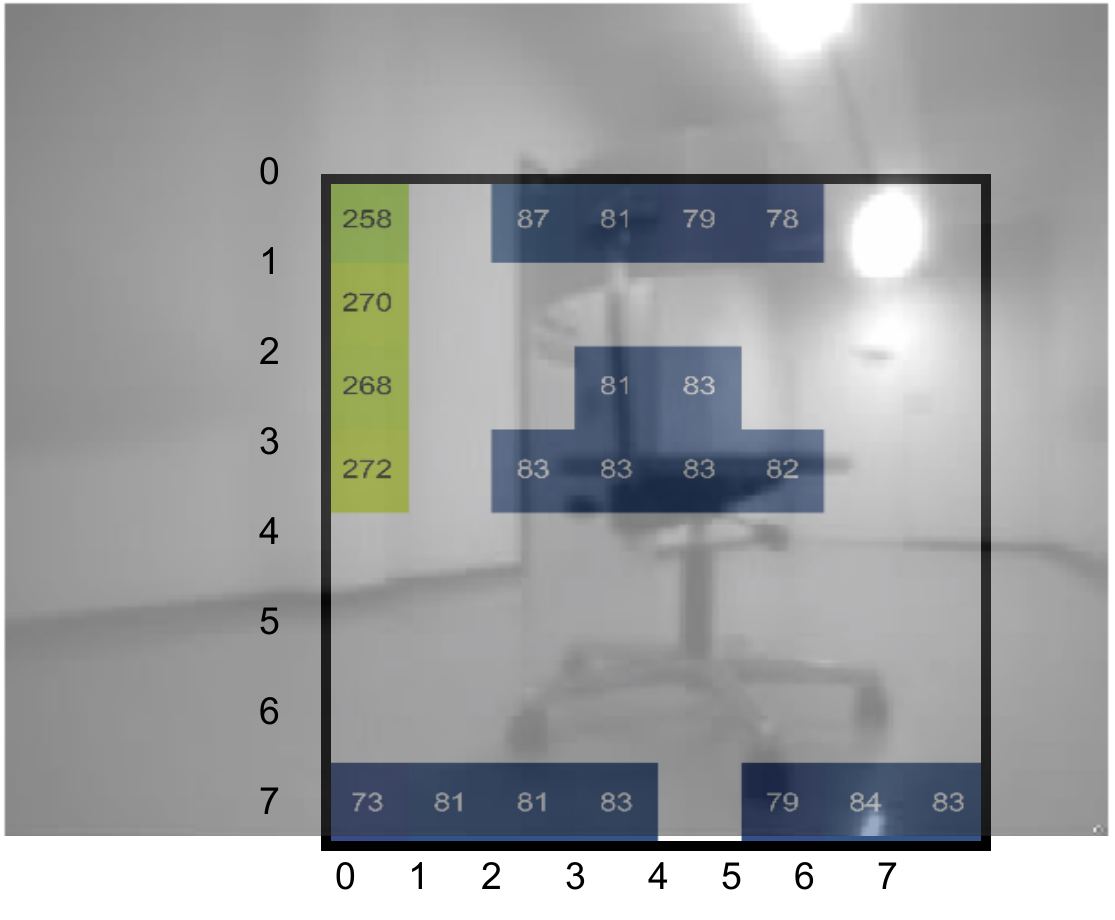}
        \caption{Pre-processed 8x8 depth map from the front-facing multi-zone ToF sensor, where invalid pixels are filtered and not shown. Each pixel features a distance expressed in centimeter and a colormap to help the reader in visualizing the depth of the front facing obstacle. The camera image is displayed as well for visualizing the FoV}
        \label{fig:dataset_chair_tof}
    \end{subfigure}
    \caption{Still scene from the \textit{O4} recording\footnoteref{note:github}. In this test, the UAV took off starting to approach an office chair placed in a large room, an human operator supervised and controlled the whole maneuver. In this specifically example, we show that at this distance from the obstacle, the multi-zone ToF sensor correctly identifies the chair sitting and the backrest, but the metallic support between the two is not fully visible. The two figures show the identical Crazyflie status from different perspective, respectively from the left to the right. (a) The grayscale camera from the \textit{AI-Deck}; (b) Pre-processed 8x8 depth map from the multi-zone ToF sensor, scene equivalent to (a). Note that the Himax HM01B0 and the VL53L5CX have a different FoV and different mounting position on the Crazyflie frame - see \Cref{fig:cf_tof_ai} for reference. The camera image is displayed as well for visualizing the FoV. }
    \label{fig:dataset_chair}
\end{figure}
For each of the 43 released records, a pair of a .csv and .dat file format are present for (\textit{a}), (\textit{b}), and (\textit{d}). Whereas, for (\textit{c}), a series of .jpg files are present, named with the acquisition frame time in milliseconds. To combine images and decode the time-series files, we also provide a Python script named "Flight\_visualizer.py", which generates a 3D visualization of the drone attitude and spatial position from the internal state estimator and the Vicon system. Moreover, images and the 8x8 ToF matrix are time-aligned and plotted together with the drone state. The script offers the possibility to test the control algorithm on the collected data. We provide an example in \textit{object\_detection} and \textit{decision\_making} functions that can be used as a reference point for future work. 
\Cref{fig:dataset_man} and \Cref{fig:dataset_chair} are respectively two representative examples from \textit{O16} and \textit{O4} recordings\footnoteref{note:github}, reporting the grayscale image and the depth matrix. In \Cref{fig:dataset_man}, the drone is hovering in a fixed position, $V_x,V_y,V_z \approx 0$ and $(yaw,pitch,roll) \approx (0,0,0)$, at \SI{1}{\meter} from the ground while a person is walking perpendicularly to the VL53L5CX FoV. In \Cref{fig:dataset_man_tof}, the 8x8 depth matrix shows the foreground distance from the nano-drone, which is reported within a centimeter precision.

Thanks to the sensor's ability to automatically detect invalid pixels, the background (out of range) is automatically subtracted, and the moving object, the foreground, is then extracted from the scene at zero-computational cost. Despite \Cref{fig:dataset_man_img} supporting the reader in understanding the test setup and the 8x8 matrix, one can already notice that the HM01B0 is saturating due to the ambient lighting. This condition could decrease the integrity of algorithms fully based on vision-based sensing. Note that the legs of the person are right at the pixel border, leading to only one pixel for them in column 4, and the person is stepping forward, leading to the right knee (pixel 5/5) being out of range.
On the other hand, \Cref{fig:dataset_chair} gives an example of a real flight controlled by a human pilot, in which the multi-zone ToF sensor does not correctly extract the object shape. The pilot took off and then swerved by \SI{180}{\degree} from the starting position and is approaching the chair at \SI{0.35}{\mps}. Despite in \Cref{fig:dataset_chair_img} an office chair is correctly visible (note that pixels in row 7 belong to the ground), the depth map in \Cref{fig:dataset_chair_tof} does not fully extract the foreground detail. Indeed, the chair sitting and backrest are identified and measured to be at \SI{83}{\centi\meter}, but the metallic support between the two is completely invisible to the multi-zone ToF sensor, which then wrongly identifies a possible safe passage between rows 1 and 2. In this scenario, the chromed and thin metallic support reflects the majority of the \SI{940}{\nano\meter} laser beam, being visible only from certain angles or at very short distances, i.e., below \SI{50}{\centi\meter}. This peculiar behavior motivates the technical choice to use an image segmentation approach instead of a pixel-granular cost function to avoid obstacles.

\section{Low-Latency Lightweight Obstacle Avoidance} \label{sec:algorithm}
\begin{figure}[t]
  \centering
  \includegraphics[width=\linewidth]{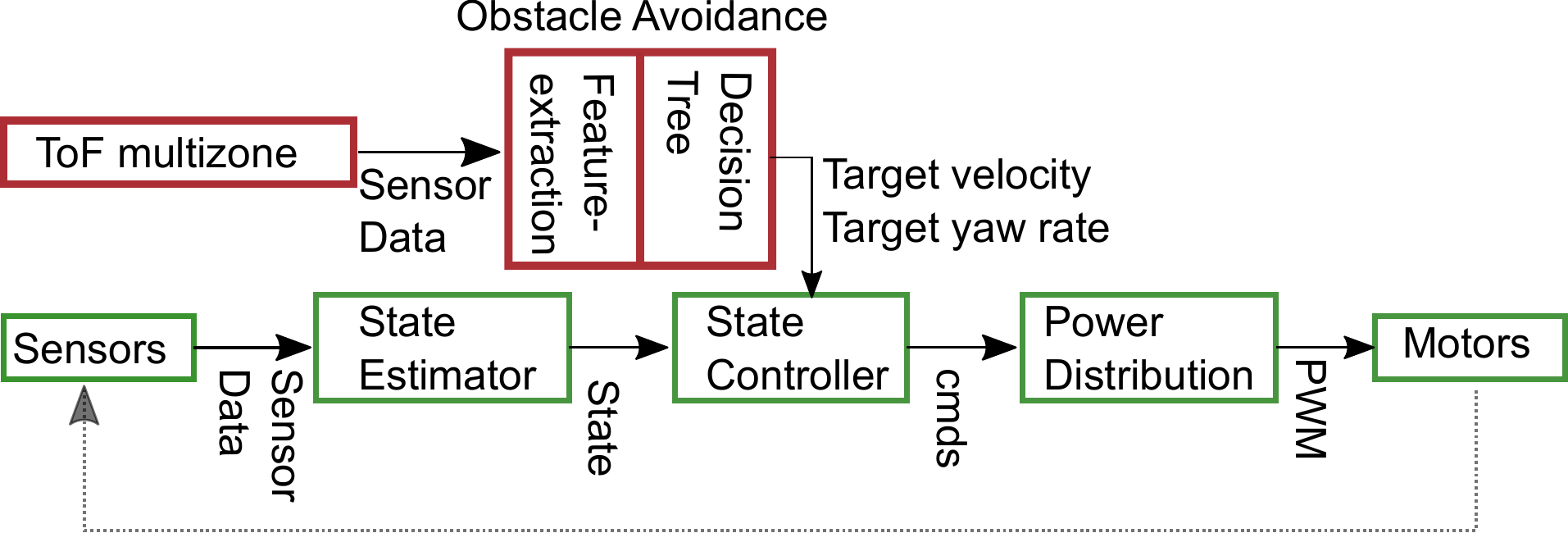}
  \caption{Integration of the obstacle avoidance algorithm into the Crazyflie control flow. Additions are shown in red, the default modules in green.}
  \label{fig:flow}
\end{figure}

This section describes the whole pipeline used to implement the obstacle avoidance onboard the Crazyflie. 
In \Cref{fig:flow} we show how the proposed algorithm is integrated with the existing open-source Crazyflie firmware. 
The blocks in green belong to the base Crazyflie firmware and are used without modification, while the blocks in red represent our contribution and implement the obstacle avoidance algorithm.
The base firmware performs state estimation relying on the information from the onboard IMU and the two sensors found on the \textit{Flow-deck v2}: the downward facing one-dimensional ToF sensor used for height estimation and the optical flow sensor used for horizontal velocity estimation.
The sensor data is fed into the eKF implemented in the base firmware, which produces the state estimate -- position, velocity, attitude.
The ``Obstacle Avoidance'' block exploits the information from the multi-zone ToF sensor, producing a forward target velocity and a steering rate that enable the drone to avoid collision with the frontal obstacles.
These commands are sent to the onboard controller implemented on the base firmware, which actuates the drone accordingly.
In our obstacle avoidance pipeline, we firstly perform feature extraction to identify the objects in the ToF frame and then use a decision tree to determine the forward velocity and steering rate.

\subsection{Feature extraction} \label{sec:feature-extraction}
After a sensor frame is obtained, the system applies a preprocessing step before running the decision tree.
Firstly we threshold the ToF frame, removing all pixels with an associated distance higher than \SI{2}{\meter} -- during the sensor characterization, we discovered that measurement validity decreases below 90\% for higher distances. 
The outcome of the thresholding step is an occupancy frame (i.e., a binary frame) which indicates the presence of the obstacles for each pixel.
In the following, the neighboring pixels are grouped in clusters that define the objects -- with this approach, the overlapping objects within the FoV are treated as one object.
We define this procedure as \textit{grouping}, and the steps of the feature extraction algorithm are presented as pseudo-code in \Cref{lst:object_clustering}.
The algorithm adds all pixels that belong to the same object to \textit{groups}. 
To mitigate the effect of noise/outliers, groups have a minimum number of 2 pixels, and single pixels (i.e., without any neighbor) are ignored. 

Our algorithm first initializes all pixels as unvisited, then passes through them one by one to check if they belong to a group. 
The grouping starts with an unvisited pixel that belongs to an obstacle, then recursively adds all neighbors with a positive occupancy status.
As we call the \textit{GroupAddDFS} function at most once per pixel (resulting in a depth-first-search), the algorithm runs in $\mathcal{O}(n)$, where $n$ is the number of pixels.  
The number of groups is in practice limited to 4. 
Several metrics characterize each group:
: (a) minimal and maximal X/Y coordinates (borders), (b) number of pixels, (c) position (averaged position of all pixels belonging to the group), (d) minimum distance to the object.




\begin{listing}[t]
\begin{minted}[mathescape=true,
             breaklines,
             escapeinside=||,
             numbersep=5pt,
             gobble=2,
             fontsize=\footnotesize,
             framesep=2mm]{python}

# for a given pixel, GroupAddDFS finds all connected, occupied, and unvisited pixels
def GroupAddDFS(pixel, group_index):
    for neighbor of pixel:
        if neighbor.occupied == true and neighbor.visited == false:
            neighbor.visited = true
            GroupAddDFS(neighbor, group_index)
            group[group_index].add(neighbor)
# finds all pixel groups/clusters
def Grouping(binary_frame):
    group_index=0
    for pixel in binary_frame:
        pixel.visited = false
    for pixel in binary_frame:
        if pixel.occupied == true:
            if pixel.visited == false:
                pixel.visited == true
                GroupAddDFS(pixel, group_index)
                if group[group_index] is not empty:
                    group[group_index].add(pixel)
                    group_index++
\end{minted}
\caption{This algorithm clusters all the occupied pixels (i.e., distance $<$ \SI{2}{\meter}) into individual groups. \textit{pixel.occupied} is a binary variable and indicates the occupancy status, while \textit{pixel.visited} indicates if the algorithm already passed through the pixel. \textit{group\_index} refers to the number of a group, and after executing the \textit{Grouping}, the value of \textit{group\_index} corresponds to the number of groups/clusters.}
\label{lst:object_clustering}
\end{listing}

\subsection{Decision tree}
\begin{figure}[t]
  \centering
  \includegraphics[width=0.8\linewidth]{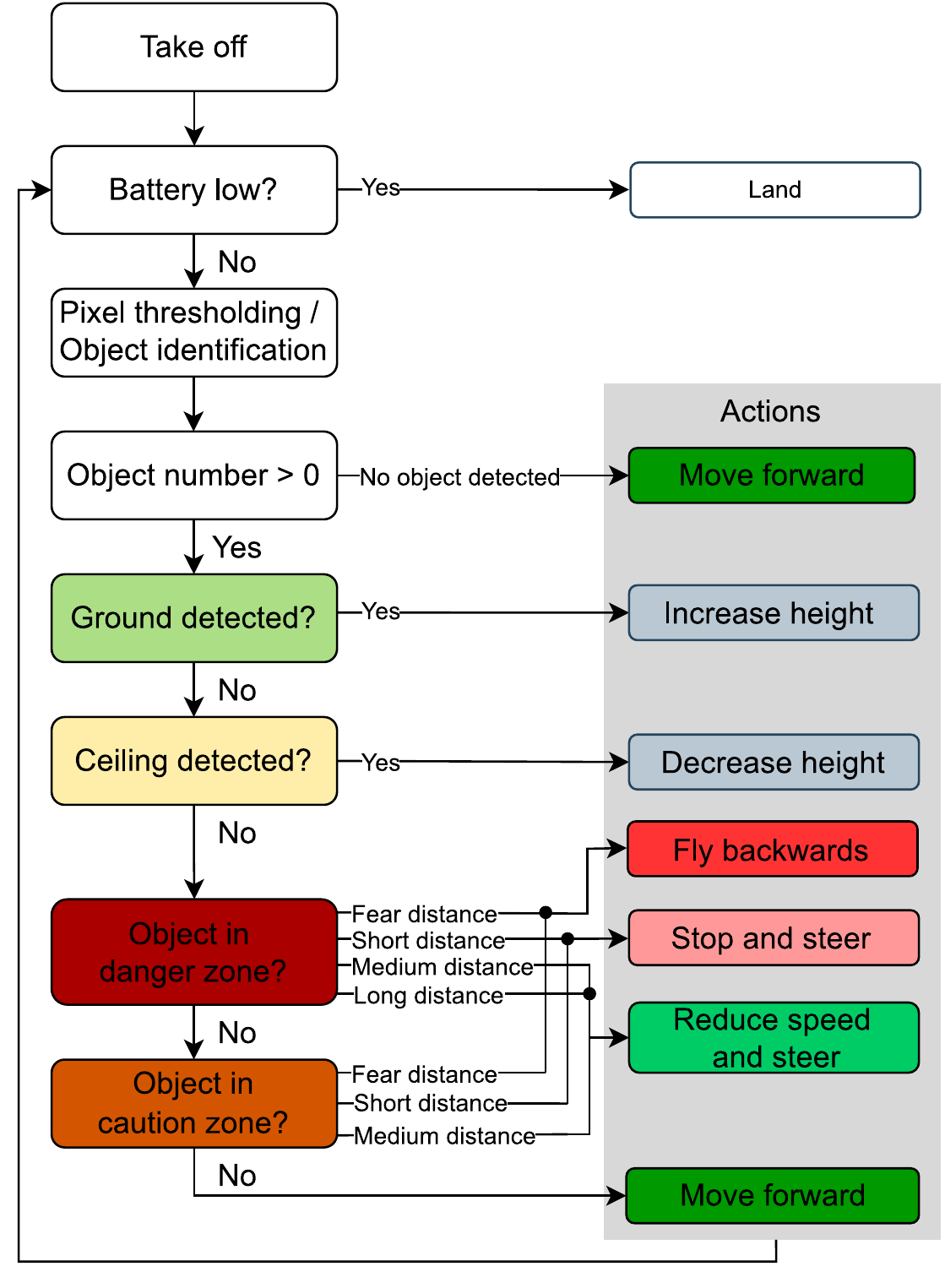}
  \caption{The obstacle avoidance flowchart, illustrating the feature extraction blocks as well as the decision tree which provides the control commands for the drone.}
  \label{fig:flow-chart}
\end{figure}

\Cref{fig:flow-chart} illustrates the flow diagram that describes how our system interprets the ToF information and generates the flying commands. 
This flow runs in a continuous loop and is designed to have low latency and low complexity as it runs using only 520800 cycles while providing accurate commands.
To ensure safety, the system constantly checks the battery level, and if the battery is low, the drone lands.
The ``Pixel thresholding / Object identification'' are related to the feature extraction presented in \Cref{sec:feature-extraction}.
Suppose the system identifies at least one object within the FoV during the feature extraction phase. In that case, the decision tree is employed as a collision avoidance algorithm to decide what flying command to apply.
The decision tree provides the flying commands (i.e., steering rate and forward velocity) based on the distance to the object and the zone where the obstacle is found within the FoV.

\begin{figure}[t]
  \centering
  \includegraphics[width=0.99\linewidth]{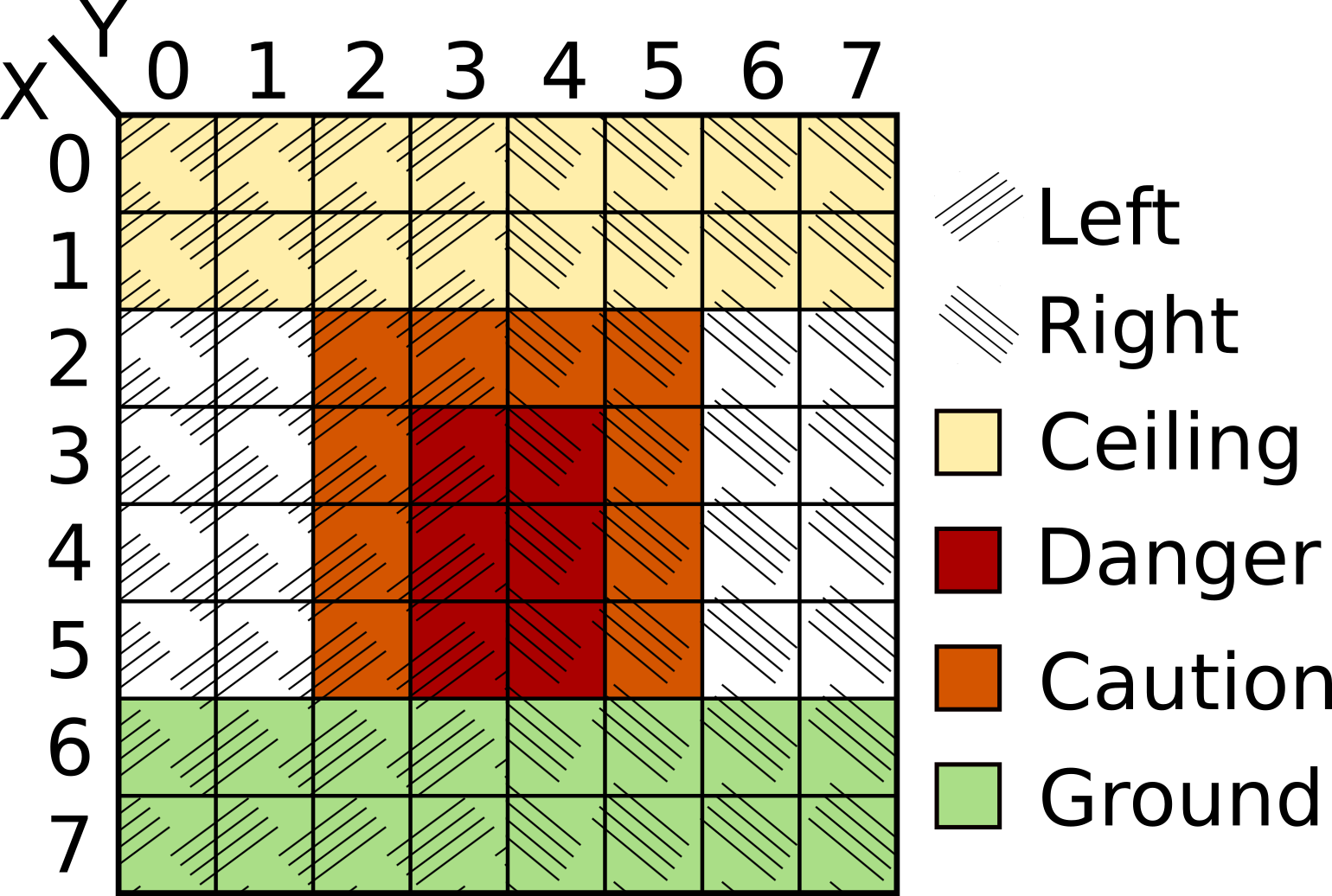}
  \caption{The multi-zone ToF sensor is configured to measure 8x8 pixels, covering a FoV of around \SI{64}{\degree} (diagonally). We define four zones, a ceiling and ground zone to not fly too close to those, as well as a danger and caution zone in the center of the FoV.}
  \label{fig:zones}
\end{figure}

The distance is split in five intervals determined by four thresholds: $d\sb{fear}$, $d\sb{short}$, $d\sb{med}$, and $d\sb{long}$ which take the values \SI{0.15}{\meter}, \SI{0.4}{\meter}, \SI{0.7}{\meter}, and \SI{1.4}{\meter}, respectively (determined empirically).
The zones are defined by dividing the FoV into four zones (i.e., ground, ceiling, caution, danger) and two sides (i.e., left and right), as shown in \Cref{fig:zones}. 
Given that we target to mainly explore indoor environments (e.g., corridors, offices), we assume that the floor and the ceiling are mostly flat. Therefore, the drone is commanded to fly at a fixed height (i.e., \SI{0.4}{\meter}) from the floor.
While the cruising height is \SI{0.4}{\meter}, the system continuously checks if the closest object is in the ceiling/ground zone, and if this is the case, it adjusts the height accordingly so that it keeps distance from the object.

If there is no obstacle in either of these two zones, the algorithm checks for the presence of the obstacle in the danger and caution zones and reacts according to the distance to the object.
For instance, if the distance to the object is smaller than $d\sb{fear}$, the drone flies backward to avoid being very close to the object -- we further motivate in \Cref{sec:results} that being close to a wall/object decreases the accuracy of the drone's state estimation.
Moreover, if the distance to the object is in the interval $(d\sb{fear}, d\sb{short})$ the drone completely stops and steers until it determines that it is safe to fly in the forward direction.
Lastly, if the distance is higher than $d\sb{med}$, the drone does not have to stop completely, but it slows down and steers while flying.

\begin{figure}[t]
  \centering
  \includegraphics[width=1\linewidth,clip]{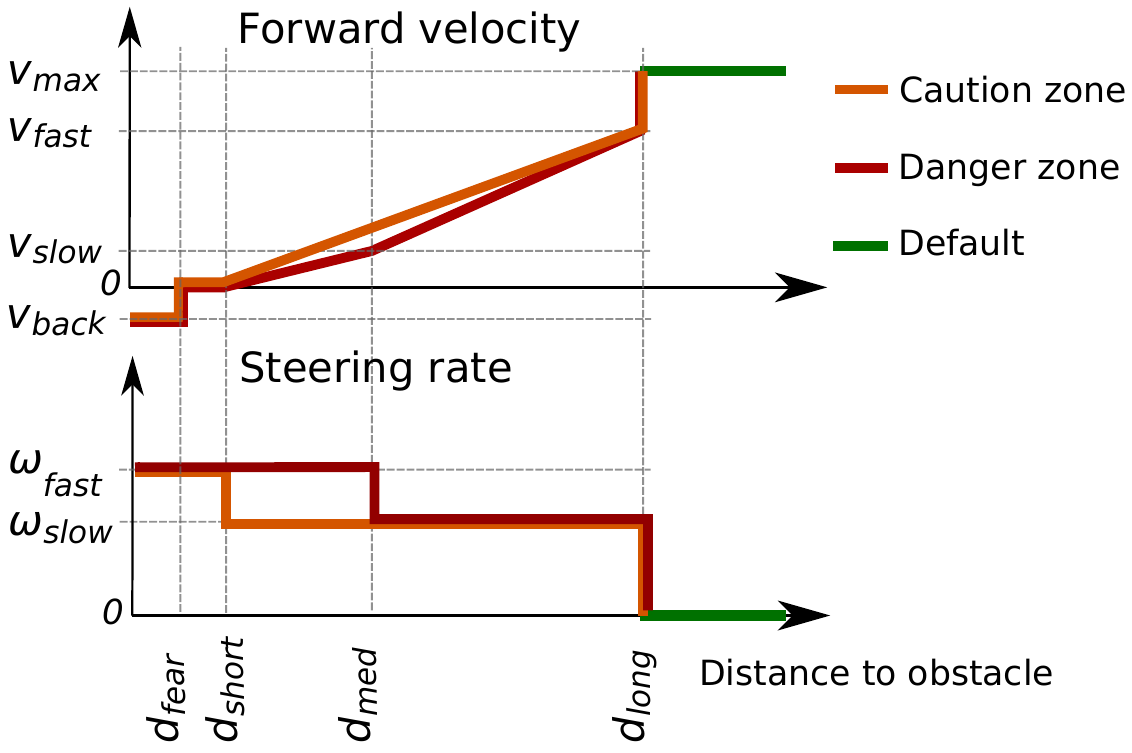}
  \caption{The curves of the commanded forward velocity and steering rate for the caution, danger, and default zones.}
  \label{fig:velocity}
\end{figure}

However, the actions presented in \Cref{fig:flow-chart} are rather simplified because the values of the velocity and steering rate depend on both zone and distance to the object.
\Cref{fig:velocity} presents the velocity and steering rate curves for the danger, caution and default (i.e., no obstacle) zones.
The $v\sb{back}$, $v\sb{slow}$, and $v\sb{fast}$ take values of \SI{-0.2}{\mps}, \SI{0.15}{\mps}, and \SI{0.85}{\mps}, respectively.
$v\sb{max}$ represents the forward velocity when no obstacle is detected and is a configurable parameter.
One can note that the forward velocity varies linearly with the distance to the object in the caution and danger zones when the distance takes values within ($d\sb{short}$, $d\sb{long}$).
However, in the danger zone, the velocity slope is slightly different; taking two possible values -- we determined empirically that this improves the system robustness.
The steering rate can take the value of either $\omega\sb{slow} = \SI{0.7}{\radian / \second}$ or $\omega\sb{slow} = \SI{1}{\radian / \second}$ as shown in \Cref{fig:velocity}.
In addition to what is shown in \Cref{fig:flow-chart}, our system also checks for dead ends.
When the drone is stuck in front of a blocked path, it would typically start oscillating left and right, trying to find the way out. 
To mitigate this issue, we implement a history mechanism that checks for these repeated oscillations and steers the drone \SI{180}{\degree} once this is detected.


\section{Results}  \label{sec:results}

This section provides a power and computational requirements analysis of our approach.
Furthermore, it presents an evaluation of our system, demonstrating the obstacle avoidance and exploration capabilities in real-world experiments. 
We evaluate the system's functionality with both static and dynamic obstacles in various environments.

\subsection{Computational load and power consumption}
Our algorithm (displayed in red in~\Cref{fig:flow}) takes 35k cycles to process one frame on average, at the maximum rate of the ToF multi-zone sensor (\SI{15}{\hertz}).
This means we add a mere 0.31\% load to the STM32F405 on the Crazyflie.
The latency from the acquired ToF image data to the flight command is \SI{210}{\micro\second} on average.
The Crazyflie control flow (displayed in green in~\Cref{fig:flow}) with a flow-deck and configured to use an eKF for state estimation needs 35\% of the computational capabilities of the STM32F405, meaning we only add a minimal additional load.

The increased power consumption from software load is hence negligible. However, the sensor typically consumes \SI{286}{\milli\watt} - if we account for the voltage regulator's efficiency, it leads to a power consumption of \SI{320}{\milli\watt}. 
Power consumption also increases because of additional weight: we add \SI{7.8}{\gram} (\SI{1.7}{\gram} flow deck, \SI{2.3}{\gram} custom deck, \SI{1.1}{\gram} heavier battery, \SI{1.3}{\gram} battery holder incl. reflective marker, \SI{1.4}{\gram} long pin headers), leading to a \SI{35}{\gram} heavy drone. The maximum payload is reached at \SI{42}{\gram}, but already with our increased weight, we see degrading maneuverability and flight time. 

\begin{figure}[t]
  \centering
  \includegraphics[width=\columnwidth]{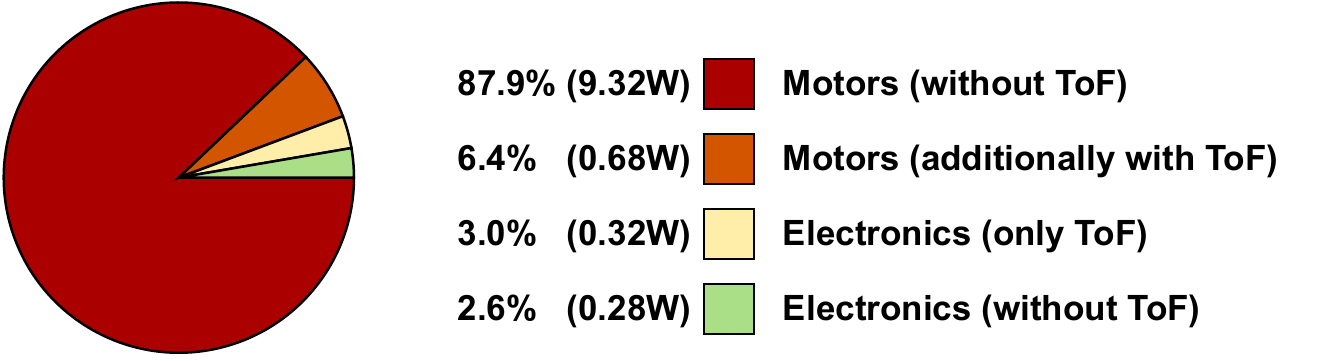}
  \caption{Power breakdown to compare the power consumption with and without our ToF deck. The ToF deck contributes only 9.4\% to the power consumption.}
  \label{fig:power_breakdown}
\end{figure}

To gain flight time and agility back, we chose a \SI{350}{\mAh} battery instead of the \SI{250}{\mAh} stock battery, at the cost of adding \SI{1.1}{\gram}. 
However, to compare the additional power load brought by our ToF multi-zone deck, we tested how long the drone can hover with and without the ToF multi-zone deck. With it, the time until the low battery warning was triggered (battery voltage measurement below \SI{3.2}{\volt} for \SI{5}{\second}) was on average 7'22'', without it 7'56''. 
Assuming the full capacity of the battery is used and assuming \SI{0.28}{\watt} for the Crazyflie electronics~\cite{mcguire2019minimal} we estimate that \SI{680}{\milli\watt} are used for carrying the additional weight of the ToF multi-zone deck and \SI{9.32}{\watt} for the remaining components. 
A power breakdown is shown in \Cref{fig:power_breakdown}. 
We see the importance of lightweight sensors - we use 9.4\% of the power for adding the multi-zone ToF sensor. However, the power needed for the sensor operation (3\%) is less than half the additional power needed to carry the shield (6.4\%).

\subsection{Static obstacle avoidance vs speed} 
\label{subsec:results-a}
In the first experiment, we evaluate the braking capability of the drone when it flies towards a \SI{1.2}{\meter} $\times$ \SI{1.3}{\meter} static obstacle made out of cardboard.
The drone flies in a straight line with a configurable maximum speed. As soon as an obstacle is detected, the drone decreases its velocity according to the algorithm presented in Section~\ref{sec:algorithm}.
We disable the steering throughout this experiment to evaluate the breaking capabilities in isolation.
The drone takes off at a distance of \SI{3.5}{\meter} far from the wall, which gives it enough space to accelerate and reach the target speed.
The evaluation is performed by sweeping the $v\sb{max}$, which is a software parameter that indicates the target velocity the drone aims to reach in the absence of any obstacle within the field of view. 

\begin{figure}[t]
  \centering
  \includegraphics[width=1.0\linewidth]{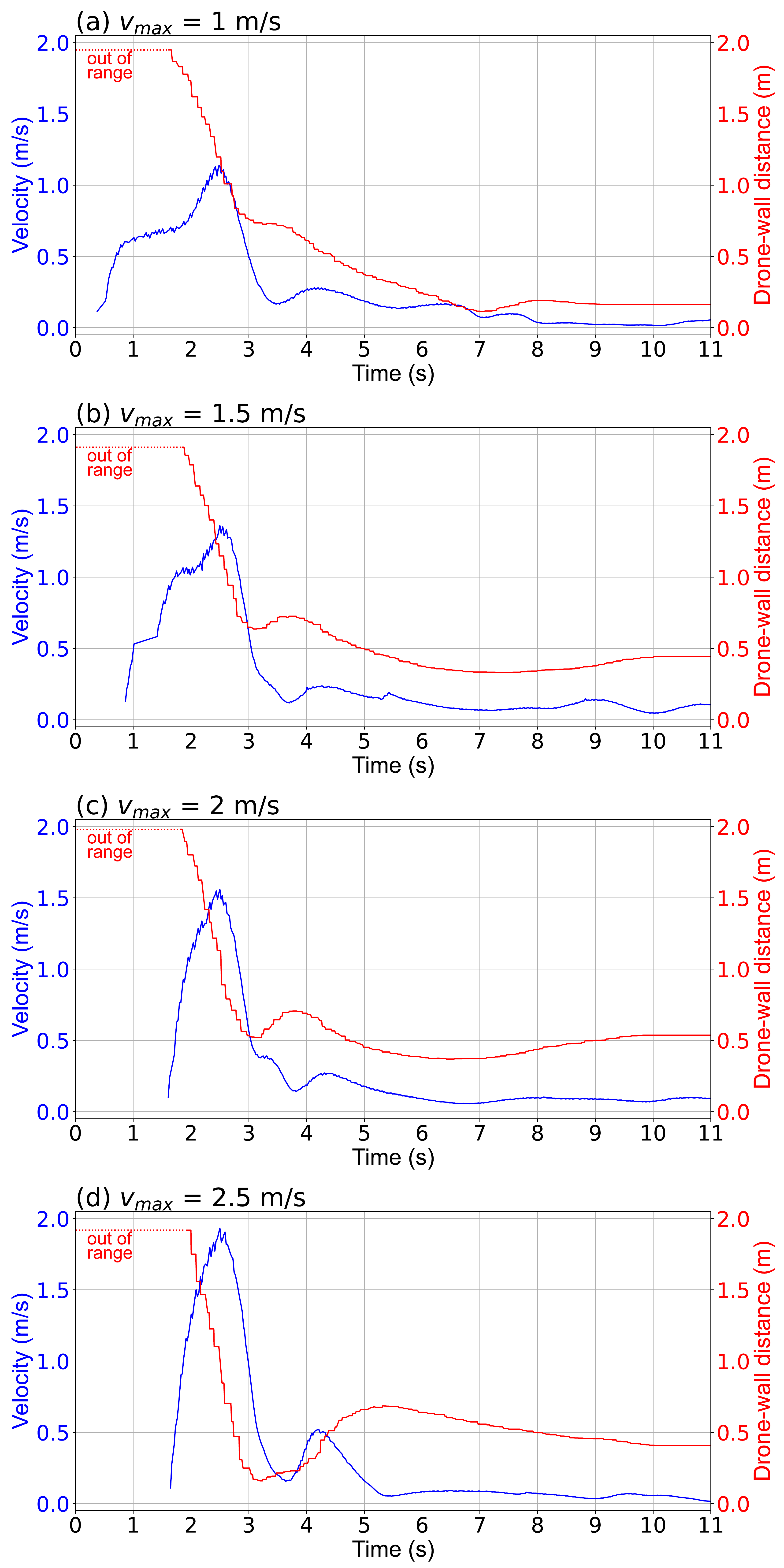}
  \caption{Braking in front of a wall with different $v\sb{max}$. In red we show the distance from the wall, measured by the onboard ToF sensor. In blue we show the velocity, recorded by the mocap system.}
  \label{fig:statc_break_wall}
\end{figure}

We perform the experiment for the following values of the parameter $v\sb{max}$: \SI{1}{\mps}, \SI{1.5}{\mps}, \SI{2}{\mps}, \SI{2.5}{\mps}, and we present the curves for drone's position and the drone -- wall distance in \Cref{fig:statc_break_wall}.
The red curve represents the distance from the drone to the wall, and the ToF sensor provides it. 
The blue curve represents the velocity of the drone, and it is logged with the Vicon at a rate of \SI{50}{\hertz}.
The subplots in \Cref{fig:statc_break_wall} were aligned in time by velocity peaks, which are \SI{1.11}{\mps}, \SI{1.34}{\mps}, \SI{1.53}{\mps}, and \SI{1.92}{\mps}, for the cases (a), (b), (c), and (d), respectively.
The drone successfully brakes in each of the four situations, and it stops about \SI{0.2}{\meter} -- \SI{0.5}{\meter} away from the obstacle.
We also point out that the braking is not very smooth due to the oscillating behavior of the velocity curve right after braking, which is visible in all situations but especially in \Cref{fig:statc_break_wall}-(d).

\begin{figure}[t]
  \centering
  \includegraphics[width=1.0\linewidth]{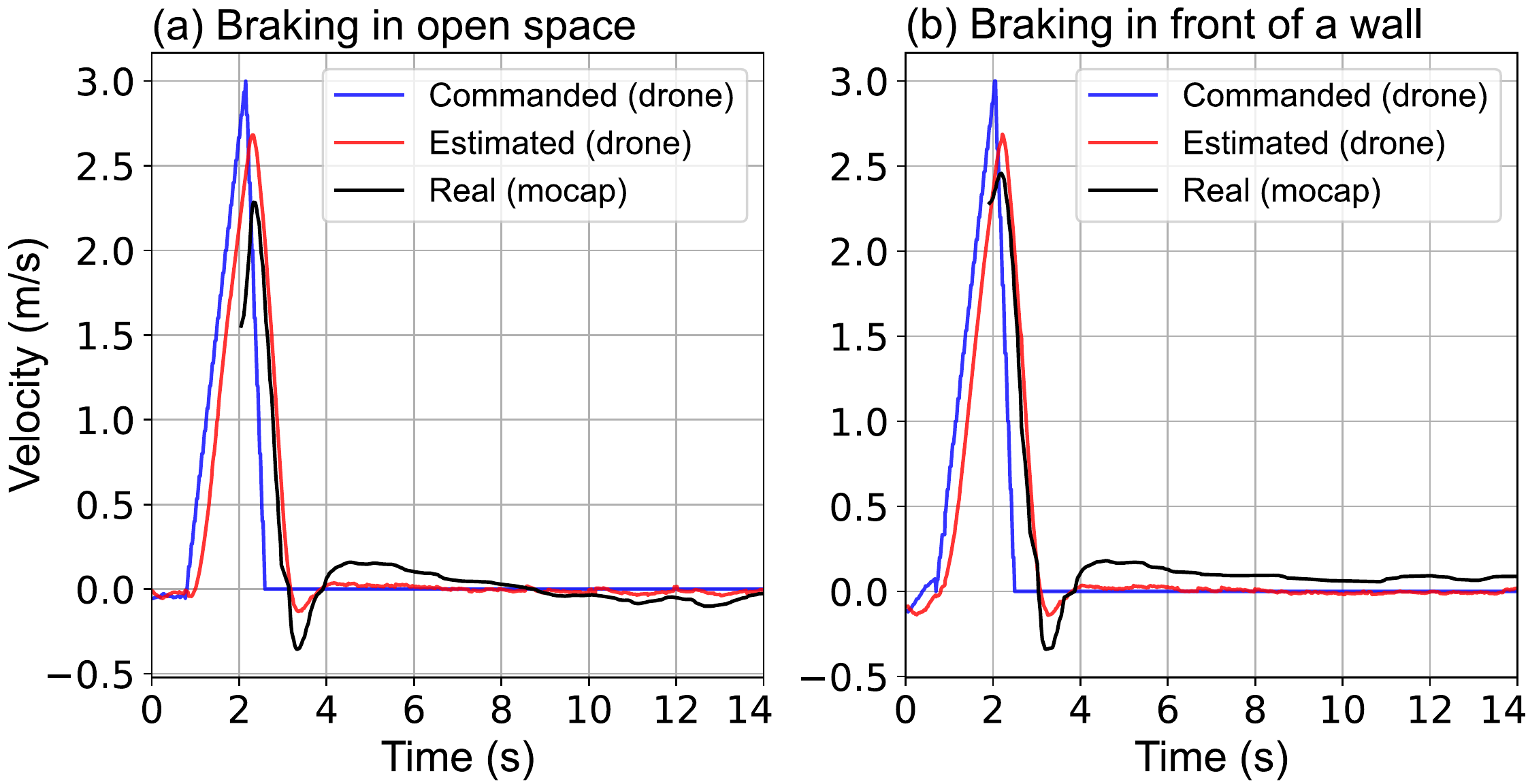}
  \caption{Comparing real, commanded and estimated forward velocities while braking in open space and in front of a wall.}
  \label{fig:brake-wall-nowall}
\end{figure}

To better investigate this effect, we perform a separate experiment where the drone is commanded to accelerate up to \SI{3}{\mps} and then suddenly brake, without using the avoidance algorithm.
Furthermore, we perform this experiment for two cases: \textit{i)} the drone flies straight in an open space with no obstacles around \textit{ii)} the drone flies straight, but a cardboard panel is mounted \SI{30}{\centi\meter} away from the stopping point of \textit{i)}.
\Cref{fig:brake-wall-nowall} shows the commanded, estimated, and actual velocities in blue, red, and black, respectively. 
While the commanded and estimated velocities are acquired from the drone directly, the actual velocity (i.e., the ground truth) is observed and logged with the Vicon system.
As its field of view is limited, the ground truth is not captured for the whole trajectory but only within the area of interest (i.e., where the drone brakes).
\Cref{fig:brake-wall-nowall}-(a) shows a velocity estimation error of about \SI{0.13}{\mps} right after the drone brakes.
Even if this error decays within about \SI{4}{\second}, it causes a forward drift as the drone believes it is stationary while it is actually moving forward.
Therefore, during aggressive braking, the sensor readings' precision decreases, impacting the drone's state estimation accuracy.
Furthermore, \Cref{fig:brake-wall-nowall}-(b) shows that this effect is exacerbated by the presence of an obstacle in the proximity of the braking point, where the decay time of the velocity estimation error is significantly longer.
This is due to wall effects that change the drone's dynamics and impact the accuracy of the down-pointing altitude sensor -- since the detection area of the altitude sensor is instead a cone than a narrow beam, staying close to walls can lead to inaccurate altitude measurements.
Therefore, poor state estimation after suddenly braking close to walls is a limitation of the drone controller itself and not of our algorithm, which explains the oscillating pattern from \Cref{fig:brake-wall-nowall}.

\subsection{Dynamic obstacle avoidance vs. speed}
\begin{figure}[t]
  \centering
  \includegraphics[width=1.0\linewidth]{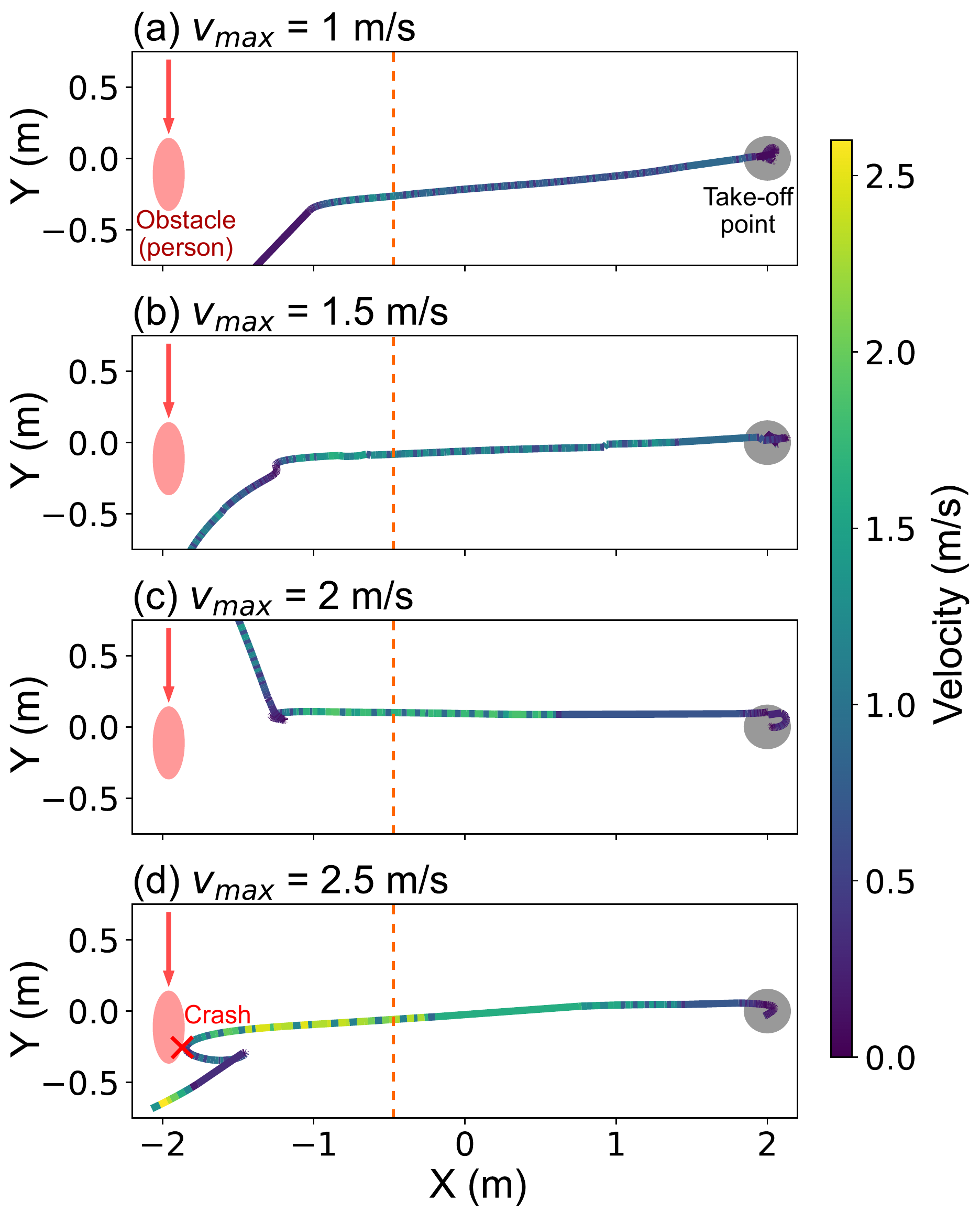}
  \caption{Brake test in front of dynamic obstacles at different velocities. A person jumps in front of the drone once it sees it at \SI{1.5}{\meter} distance. The red arrow shows the direction the person comes from, the red ellipse where it jumps to. The drone takes off in the grey circle and is headed in negative X-direction.}
  \label{fig:dynamic_brake_pic}
\end{figure}

One of the key features of an indoor autonomous drone is the ability to avoid unpredictable dynamic obstacles, especially moving persons.
Therefore, in the following experiment, we assess the avoidance capability when a person unexpectedly steps in front of the drone, leaving about \SI{1.5}{\meter} for braking and collision avoidance. 
Similarly to the experiment in Section~\ref{subsec:results-a}, we sweep the velocity in the range \SI{1}{\mps}, -- \SI{2.5}{\mps} with a step of \SI{0.5}{\mps} and report the results in \Cref{fig:dynamic_brake_pic}.
Each subplot shows the drone's trajectory, color-coded by its velocity.
The red arrow indicates the moving direction of the person, while the orange dashed line indicates the drone's position when the person jumped in front of it.
The drone's trajectory and velocity were acquired with the mocap, which observed the peak velocities of \SI{1.36}{\mps}, \SI{1.65}{\mps}, \SI{1.93}{\mps}, and \SI{2.66}{\mps} for the cases (a), (b), (c), and (d), respectively.
The experiments in \Cref{fig:dynamic_brake_pic}-(a)--(c) show successful collision avoidance, and at low velocity (i.e., \Cref{fig:dynamic_brake_pic}-(a)), the avoidance appears to be smoother because the drone has more time to react.
In the experiment depicted in \Cref{fig:dynamic_brake_pic}-(d), the drone does not manage to brake within \SI{1.5}{\meter}, and it collides with the person.
To ensure the reliability of the experiments, we performed several trials for each value of $v\sb{max}$ and observed very similar behaviors to the ones presented in \Cref{fig:dynamic_brake_pic}.
However, for $v\sb{max}=\SI{2.5}{\mps}$, the drone does not always crash because of the collision but also because it gets unstable during the sudden braking.

\begin{figure}[t]
  \centering
  \includegraphics[width=1.0\linewidth]{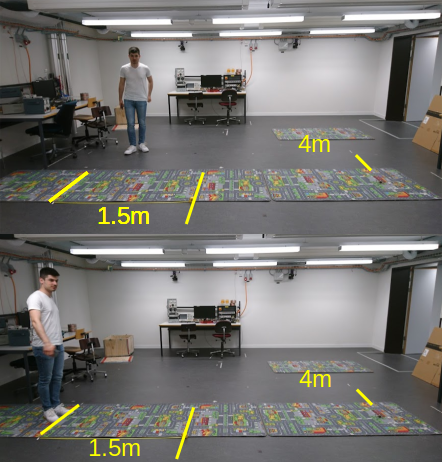}
  \caption{Our test setup for testing dynamic braking. A person jumps in front of the drone once it sees it at \SI{1.5}{\meter} distance.}
  \label{fig:dynamic_brake_pic_vlad}
\end{figure}

\subsection{Narrow pipe}
\label{subsec:results-pipe}
\begin{figure}
  \centering
  \includegraphics[width=1.0\linewidth]{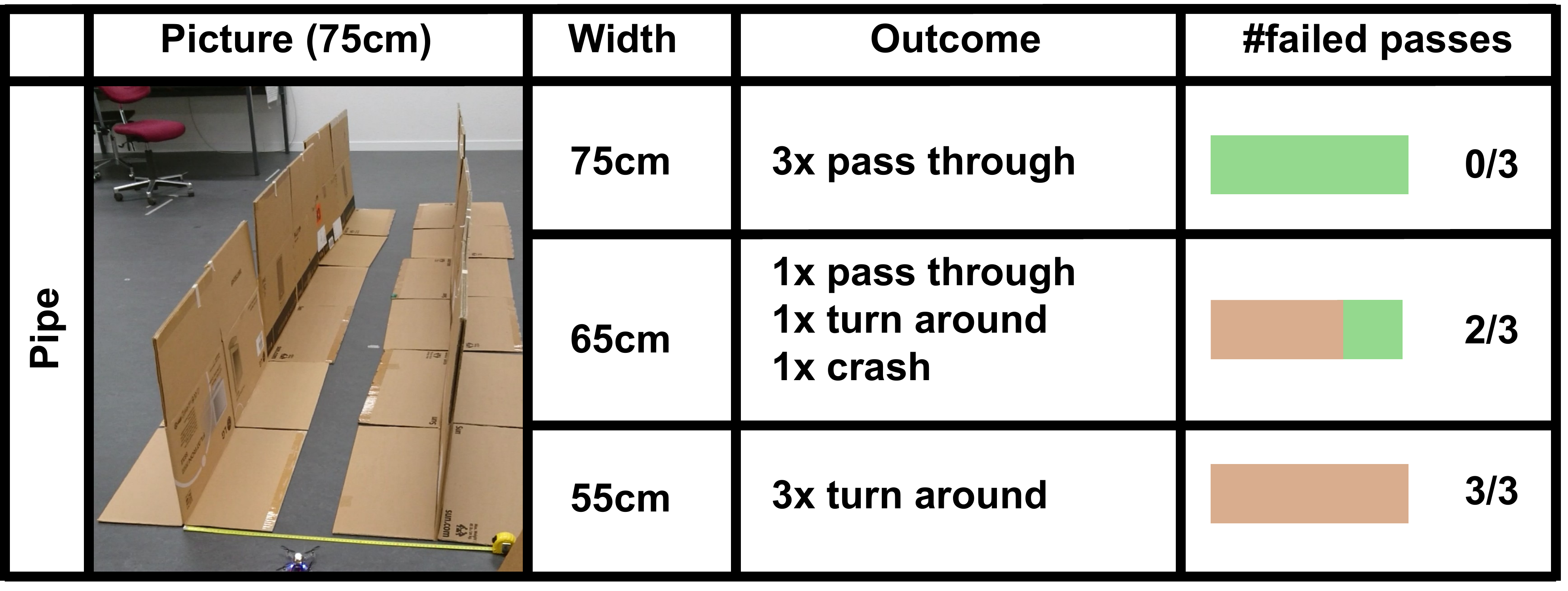}
  \caption{Flying through a narrow pipe at different widths.}
  \label{fig:pipe}
\end{figure}

For testing the drone's capability to explore narrow spaces, we built \SI{4}{\meter} long pipes of varying widths. We started the drone at the beginning of the pipe, facing in the desired flying direction (heading straight through the pipe). We did three trials at each width - at \SI{75}{\centi\meter} width, the drone always passed through without any issue; at \SI{65}{\centi\meter}, only one of the trials was successful, and at \SI{55}{\centi\meter}, the drone did not even enter the pipe once. Note that the drone can pass through much smaller gaps if they are not pipes but shorter obstacles, such as passing underneath a chair, as in Section~\ref{subsec:results-reliability}. Following the geometric relations shown in \Cref{fig:droneangle}, we can compute that the danger zone introduced in section~\ref{sec:algorithm} is \SI{33}{\centi\meter} wide at the reaction distance (\SI{1.4}{\meter}); however, the caution zone is \SI{66}{\centi\meter} wide. As shown in \Cref{fig:flow-chart}, obstacles in the caution zone will already cause the drone to turn; however, slowly enough to successfully pass through the pipe. \Cref{fig:pipe} shows the pipe and the results.

\subsection{Flying in a room with static obstacles}
\label{sec:maze}
\begin{figure}[t]
    \centering
        \centering
        \includegraphics[width=\columnwidth, trim={0 0 0 0},clip]{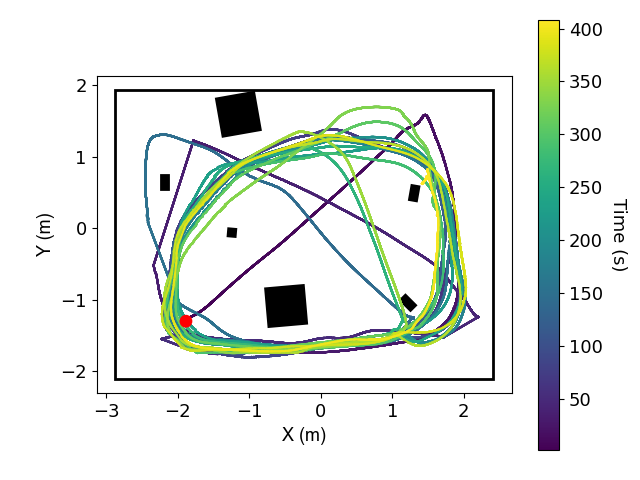}
        \caption{Flying for 6 minutes and 45 seconds in an environment with static obstacles. The color map indicates the time-course, while the data is logged with the Vicon system. The drone flies at a target height of \SI{0.5}{\meter}, with a maximum acceleration of \SI{1.5}{\meter / \second\squared}. The flight path converges to the same loop because the algorithm is deterministic.}
        \label{fig:staticmap}
\end{figure}

For this test, we built a closed environment in which we can track the drone with the Vicon system. We built a cardboard maze, as shown in the top row in \Cref{fig:flighttimes}. All obstacles are between \SI{0.6}{\meter} and \SI{0.8}{\meter} high. To verify the reliability of our system, we started the drone at different random take-off points in the maze. We set the maximum target velocity to \SI{1}{\mps} in all tests, as the environment is so cluttered that the drone always sees obstacles and is in the fixed velocity region anyway (see Section~\ref{sec:algorithm}). In \Cref{fig:staticmap}, we show one example of the flight path. We repeated the experiment 3 times without crash, with flight times of 6'31'', 6'34'' and 6'45''. The drone flies at a target height of \SI{0.5}{\meter}, with a maximum acceleration of \SI{1.5}{\meter / \second\squared} and a minimum acceleration of \SI{-20}{\meter / \second\squared}. As our algorithm is deterministic, the flight path converges to almost always the same loop. On average, we covered \SI{206}{\meter} during the flights, resulting in an average velocity of \SI{0.52}{\mps}. 

\subsection{Reliability test}
\label{subsec:results-reliability}
\begin{figure}[t]
  \centering
  \includegraphics[width=1.0\linewidth]{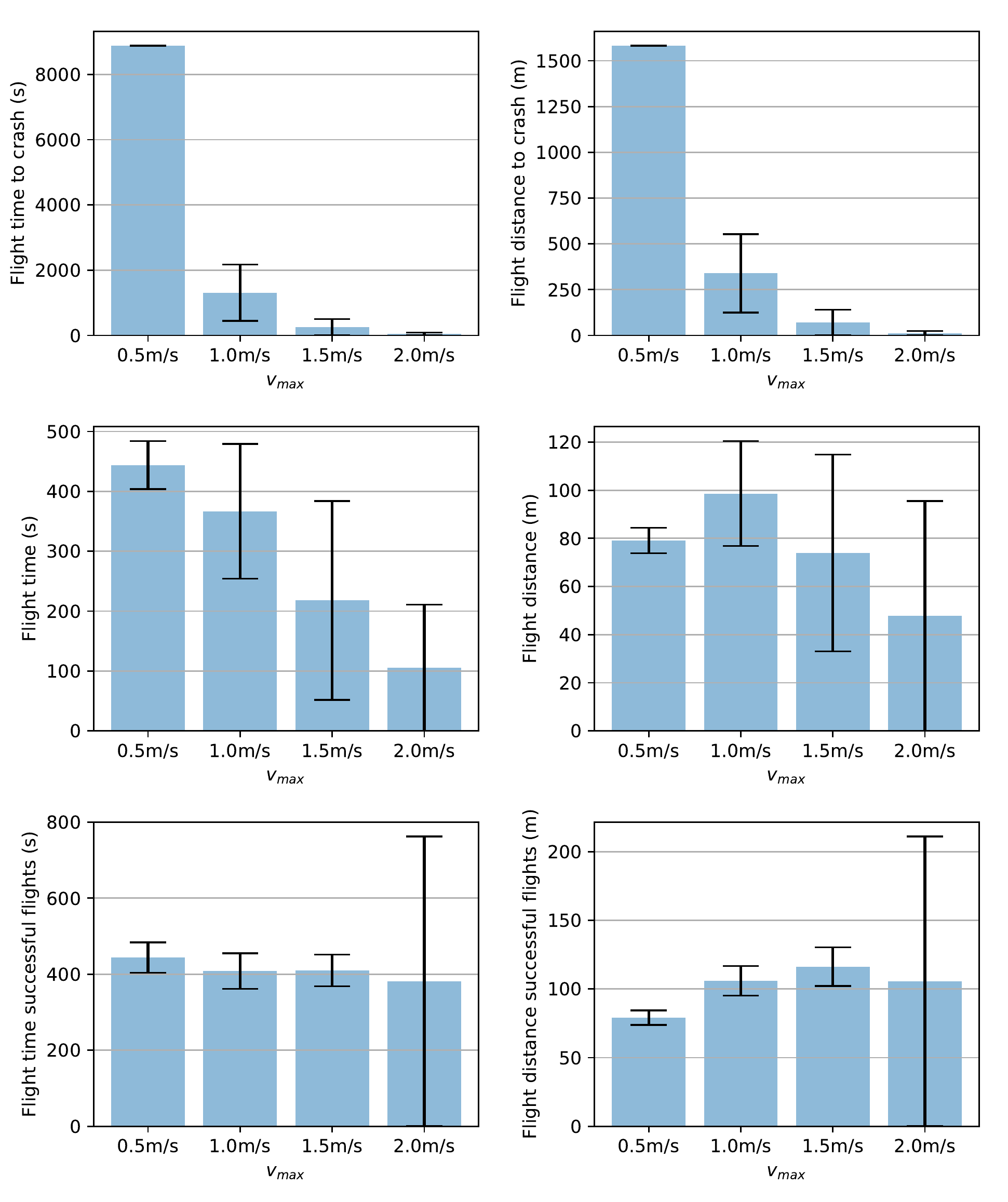}
  \caption{In the top row, we show the average time and distance until we experience a crash - not counting battery changes besides the end of the test to display this metric also when we have 100\% reliability. In the middle row, the average flight time and distance are shown. In the bottom row, we also display the average flight time and distance but only consider successful flights. The blue bar represents the average, and the black line the variance.}
  \label{fig:reliability_plot}
\end{figure}

\begin{figure}[t]
  \centering
  \includegraphics[width=1.0\linewidth]{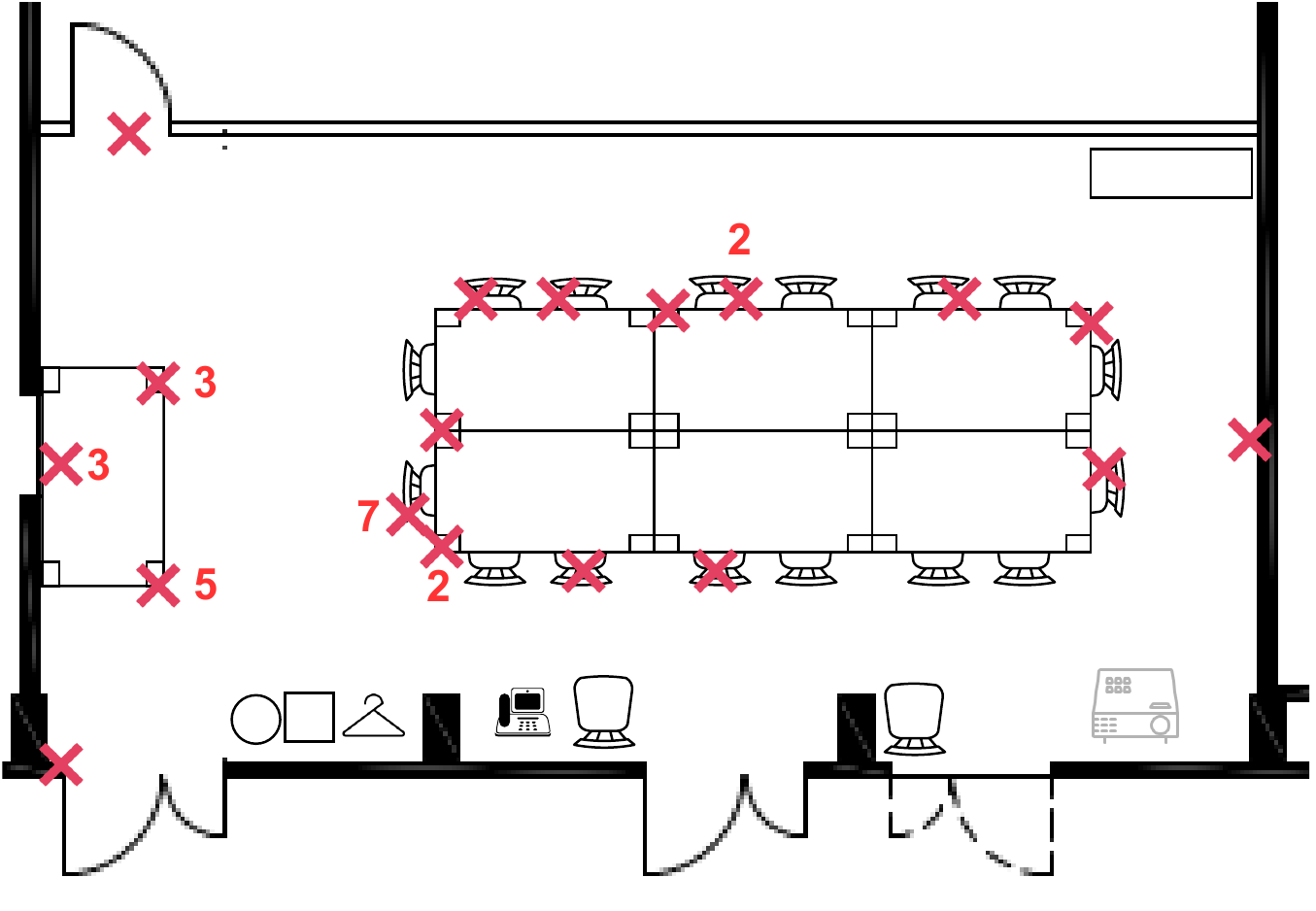}
  \caption{Floorplan of the meeting room in which we ran the reliability test. Red crosses show the recorded crashes.}
  \label{fig:floorplan}
\end{figure}

Reliability in a real-world scenario was assessed by 20 flights each at 4 different maximum target velocities in an office environment, more precisely an \SI{11}{\meter}$\times$\SI{6}{\meter} meeting room. The floorplan of the environment can be found in \Cref{fig:floorplan}. The meeting room features 7 tables, chairs, other utilities like a projector, phone, jacket rack, and several usually closed doors. Occasionally, people pass through the meeting room. We configured the drone to fly at \SI{0.4}{\meter} over the ground and tested it with four different maximum velocities - \SI{0.5}{\mps}, \SI{1}{\mps}, \SI{1.5}{\mps} and \SI{2}{\mps}. At \SI{0.5}{\mps}, we did not experience any crashes in 20 flights but always landed safely after a low battery warning. At higher velocities, the reliability dropped to 80\% at \SI{1}{\mps} and even 40\% resp. 10\% at \SI{1.5}{\mps} and \SI{2}{\mps}. Note that even when experiencing a crash, the drone often completed many successful obstacle avoidance scenarios beforehand, as one trial is not one obstacle avoidance scenario but several minutes of fully autonomous flight. We log the internal state estimation as an additional measure  and compute the covered distance. We display our results in \Cref{fig:reliability_plot}. We see that the maximum velocity only weakly influences the flight time for successful flights. The distance covered counting only successful flights is maximized at \SI{1.5}{\mps}, but note that at \SI{2}{\mps}, only 2 out of 20 flights were successful, leading to high variance. Looking at all flights, we observe the maximum of the covered distance at \SI{1}{\mps}, even though the average flight time is higher at \SI{0.5}{\mps}. We conclude that we can only fly at high speeds in easy environments (big, non-reflective obstacles). For office environments, \SI{1}{\mps} covers most distance per flight, but \SI{0.5}{\mps} is more reliable. 

\subsection{Different environments at different speeds}
\begin{figure}
  \centering
  \includegraphics[width=1.0\linewidth]{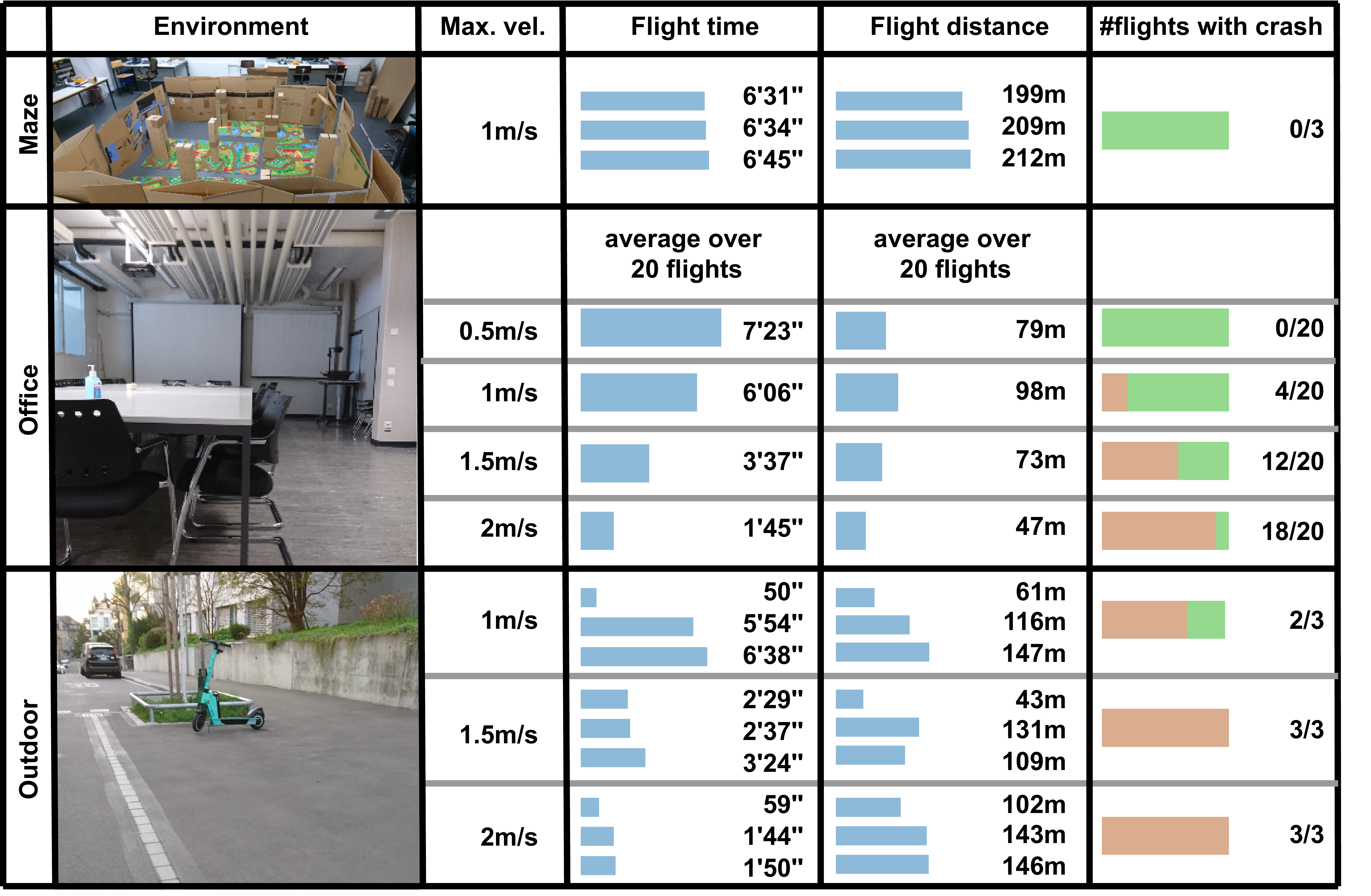}
  \caption{Real-world assessment of the proposed perception system in a controlled environment \textit{Maze}, in a general office room \textit{Office} and in \textit{Outdoor}. Results are reported for each velocity, comparing the flight time, the traveled distance, and the reliability, expressed in the number of tests that include at least one crash.}
  \label{fig:flighttimes}
\end{figure}

We tested the limits of our system by performing fully autonomous flights in various challenging environments at different maximum target velocities.
In Section~\ref{sec:maze}, we already described our baseline test - flying in a cardboard maze. Those obstacles all have the same non-reflective surface, are all taller than how high the drone will fly, and the ground is flat. In this environment, we can fly autonomously for on average 6.5 minutes with our system until a landing procedure is automatically triggered because of a low battery. In this scenario, we always see obstacles, so we do not accelerate over \SI{1}{\mps} even if we would allow it and hence only tested at this maximum target velocity. 
To challenge the drone in real-world environments, we also tested in an office environment, to be more specific, a large meeting room, described in Section~\ref{subsec:results-reliability}, and outdoors. Those environments are much more challenging, as they feature objects of various forms and surfaces. In \Cref{fig:flighttimes} we show the average flight times, covered distance and crash reasons at different maximum target velocities. 

In general, we observe that, as expected, a slower maximum target velocity leads to fewer crashes. Almost all crashes are due to highly reflective obstacles, such as metal chair legs or cars. While approaching reflective obstacles frontally at a rather low speed can work, as the sensor will measure the reflected light, approaching them at steep angles leads to failure in recognizing them. This is due to almost no deflection and hence no light coming back being sensed by the sensor. Also, small obstacles can only be detected from shorter distances, as then more light is deflected. This leads to more crashes at higher velocities, as we need more time to brake. We also observe the highest covered distance in the maze - even though the drone almost always sees obstacles and thus rarely flies at high speeds, there are no narrow dead-ends and no obstacles requiring height adjustments due to exclusively large obstacles. The office environment is far more complex, featuring tables and chairs between which it is challenging and slow to find an obstacle-free path. 

We also tested the drone outdoor in a hilly environment, but as the distance is computed from the internal state estimation, we can not take the climb into account. 
Over all different environments and speeds, we can say that for the highest reliability, the maximum target speed should be set to \SI{0.5}{\mps}, unless in environments with large and non-reflective obstacles, \SI{1}{\mps} is also possible. The flight distances are strongly influenced by the time the drone spends slowing down because of obstacles and getting out of dead ends. In general, high reliability is also beneficial for maximizing the flight distance. However, flight speeds up to \SI{1}{\mps} can lead to longer covered distances, especially in environments with big obstacles, such as the maze.


\section{Discussion and Future Work}
The key element of our system is a lightweight and reliable obstacle avoidance algorithm that leaves enough resources (computationally and energy-related) for other tasks. 
Thus, we foresee our work as a base for many future applications since reliable obstacle avoidance is only the first task towards accomplishing autonomous flying. 
This section provides an open discussion about the future work that can extend our perception system: either by integrating additional hardware (i.e., sensors) or by using more advanced data processing techniques. 

The primary constraint in adding more sensors comes from the additional weight, so we aimed to design the custom ToF deck as light as possible, leaving enough weight budget for additional sensors. 
We showed that it is possible to fly with both the multi-zone ToF deck and the AI-deck during the dataset acquisition.
Developing a sensor fusion algorithm that can use the camera and ToF sensor information could improve the obstacle avoidance robustness and enable new capabilities, such as reliable object recognition.
Adding a rear and side-facing ToF sensors on our custom deck would extend the overall FoV and, therefore, the system awareness, enabling the system to avoid rear/side approaching dynamic obstacles.
One of the main limitations of our system is dealing with highly reflective materials from extreme angles.
Even if such material also poses challenges for traditional cameras, novel miniature radars have the potential to mitigate these issues and complement the ToF sensors.

Incorporating more sensors would result in more information to be processed and, therefore, an increased need for computational resources.
Our algorithm works with a relatively low dimensional input (i.e., 64-pixel depth map) and requires about $0.31\%$ load from the STM32F405 microcontroller on board the Crazyflie.
This not only leaves a large computational budget for developing more complex algorithms, but also enables the system to deal with larger dimensional inputs.
The multi-zone ToF sensor released by STMicroelectronics is the first of its kind in terms of precision, form factor and pixel number.
However, further releases could come with improved performance, such as a higher measurement range, number of pixels or data rate.
Since higher dimensional outputs would not be as straightforward to process and interpret as in our case (i.e., 8x8), more complex algorithms such as CNNs could be a good candidate for dealing with a larger input and therefore enabling new functionalities, such as flying in uneven terrains (e.g., stairs) or detecting narrow passages.
Even if CNNs usually require large amounts of memory and computational resource, novel parallel system-on-chips -- such as the GAP8 (PULP-based) on-board the AI-deck -- proved to be very effective in running such models~\cite{niculescu2021improving} given the real-time constraints of autonomous navigation.

\section{Conclusion}
The paper presented an on-board obstacle avoidance perception system to enable autonomous navigation with nano-UAVs. It allows nano-UAWs to autonomously explore office environments reliably, only using on-board computing. We used a Crazyflie 2.1 that already features an IMU, extended by a flow deck and our multi-zone ToF deck, featuring a forward-facing 64 pixels ranger sensor. All our processing is done on-board, easily fitting on an STM32F405 microcontroller next to the flight controller, only using up $0.31\%$ of the computational power and featuring a \SI{210}{\micro\second} latency. The power to lift the additional sensor with all accompanying electronics as well as the supply of it totals in less than $10\%$ of the whole drone, making a flight time of around 7 minutes possible. We tested our system in various challenging environments, achieving autonomous flights with distances up to \SI{212}{\meter}. The 100\% reliability and high agility at a low speed in an office environment provide a base for many more complex future applications. We also provide a dataset with ToF, state estimation and camera data to learn or simulate future applications.

\section*{Acknowledgments}
The authors thank STMicroelectronics for the support provided during the development of this work. Moreover, this work was partially supported by Politecnico di Torino outgoing mobility program and EDISU international mobility grant. Thanks to Iman Ostovar for his work and professor Ernesto Sanchez for his guidance and support. The authors would also like to thank \textit{armasuisse Science \& Technology} for partially funding this research.


 
%

\bibliographystyle{IEEEtran}
\bibliography{bibliography}



\end{document}